\documentclass[10pt,twocolumn,letterpaper]{article}

\usepackage[pagenumbers]{iccv} 

\usepackage{xspace}
\usepackage{multirow}

\usepackage{algorithm}
\usepackage{algpseudocode}

\usepackage{amssymb}
\usepackage{pifont}
\newcommand{\cmark}{\ding{51}}%
\newcommand{\xmark}{\ding{55}}%

\newcommand{\shortdatasetname}{\textsc{ITD}\xspace}
\newcommand{\datasetname}{Inherent Temporal Dependencies\xspace}

\newcommand{\shortadvmeminitname}{\textsc{AdvMem}\xspace}

\usepackage{graphicx}
\usepackage{amsmath}
\usepackage{float}

\usepackage{subcaption}

\usepackage[dvipsnames]{xcolor} 

\usepackage{booktabs}
\usepackage{tabularx}
\usepackage{soul}
\usepackage{comment}

\makeatletter
\DeclareRobustCommand\onedot{\futurelet\@let@token\@onedot}
\def\@onedot{\ifx\@let@token.\else.\null\fi\xspace}
\def\eg{\textit{e.g}\onedot\xspace}
\def\ie{\textit{i.e}\onedot\xspace}
\def\etal{\textit{et al.}~}

\makeatother

\definecolor{iccvblue}{rgb}{0.21,0.49,0.74}
\usepackage[pagebackref,breaklinks,colorlinks,allcolors=iccvblue]{hyperref}

\title{\shortadvmeminitname: Adversarial Memory Initialization for Realistic Test-Time Adaptation via Tracklet-Based Benchmarking}

\author{Shyma Alhuwaider\thanks{Corresponding Author:  \texttt{shyma.alhuwaider@kaust.edu.sa}} \quad Motasem Alfarra \quad Juan C. Pérez \quad  Merey Ramazanova \quad Bernard Ghanem\\
Center of Excellence in Generative AI, KAUST, Saudi Arabia  \\
}

\begin{document}
\maketitle

\begin{abstract}
\label{sec:abstract}
We introduce a novel tracklet-based dataset for benchmarking test-time adaptation (TTA) methods. 
The aim of this dataset is to mimic the intricate challenges encountered in real-world environments such as images captured by hand-held cameras, self-driving cars, \textit{etc}. 
The current benchmarks for TTA focus on how models face distribution shifts, when deployed,  and on violations to the customary independent-and-identically-distributed (\iid) assumption in machine learning. 
Yet, these benchmarks fail to faithfully represent realistic scenarios that naturally display temporal dependencies, such as how consecutive frames from a video stream likely show the same object across time. 
We address this shortcoming of current datasets by proposing a novel TTA benchmark we call the ``\datasetname''~(\shortdatasetname) dataset. 
We ensure the instances in \shortdatasetname~naturally embody temporal dependencies by collecting them from tracklets---sequences of object-centric images we compile from the bounding boxes of an object-tracking dataset. 
We use \shortdatasetname~to conduct a thorough experimental analysis of current TTA methods, and shed light on the limitations of these methods when faced with the challenges of temporal dependencies. 
Moreover, we build upon these insights and propose a novel adversarial memory initialization strategy to improve memory-based TTA methods.
We find this strategy substantially boosts the performance of various methods on our challenging benchmark.
\footnote{Code: \href{https://github.com/Shay9000/advMem.git}{github/Shay9000/advMem.git}}.
\end{abstract}
\vspace{-0.25cm}

\section{Introduction}
\label{introduction}

Deep neural networks (DNNs) have demonstrated impressive performance across various domains~\cite{he2016deep}. However, their reliability often diminishes in real-world scenarios due to natural corruptions and distribution shifts~\cite{imagenetr,imagenetc,3dcc}. These shifts can manifest as unforeseen distortions that cause the input data to deviate from the model’s training distribution. Additionally, the distribution of image classes may differ from what the model has learned, further compounding the challenge.
Consider images captured by hand-held cameras—these introduce two major difficulties: (1)~the visual distribution may differ significantly from training data, such as in foggy or rainy conditions, and (2)~images arrive sequentially as part of a continuous video stream. The first issue represents a distribution shift in the visual space, while the second introduces temporal dependencies that break the \textit{independence} assumption inherent in conventional training (\iid). Addressing both aspects is crucial for ensuring the robustness of DNNs in practical deployments.

\begin{figure*}[t]
    \centering
    \includegraphics[width=0.8\textwidth]{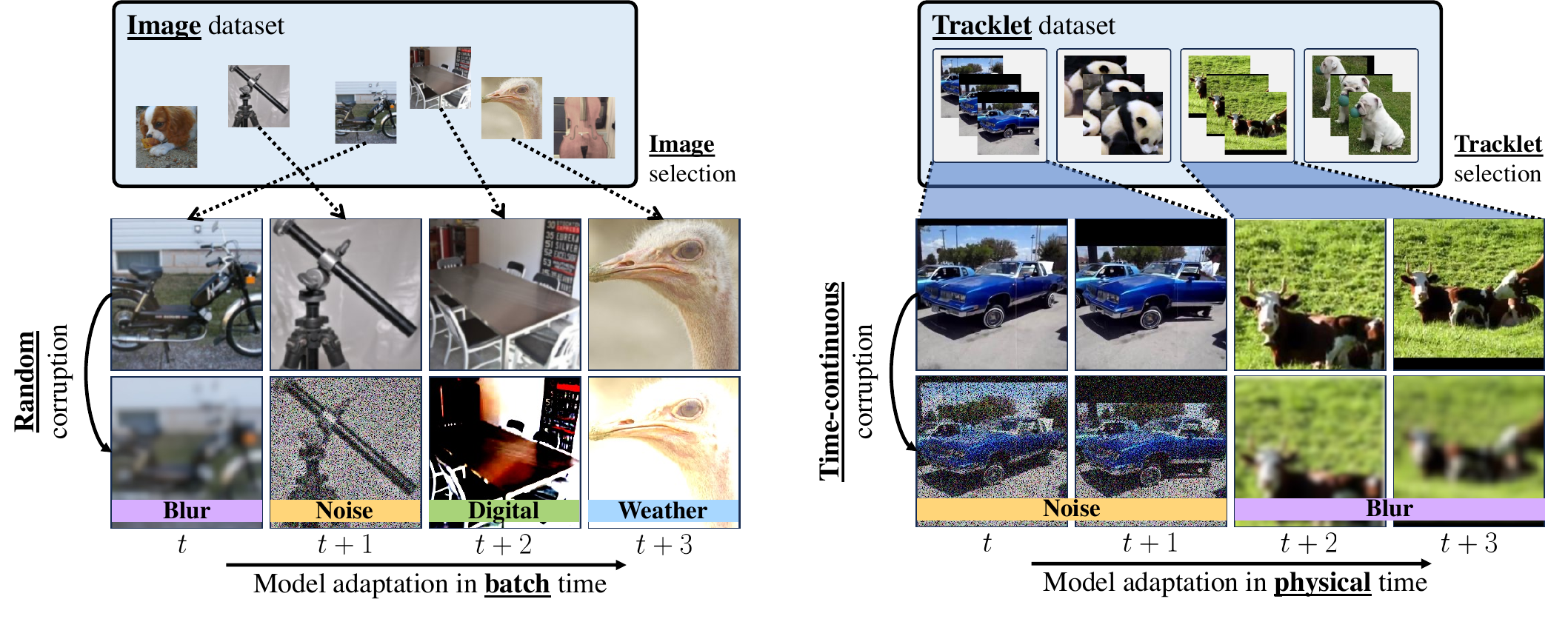}
    \caption{
    \textbf{A tracklet-based benchmark for realistic evaluation of Test-Time Adaptation (TTA) methods (\datasetname).} 
    \textit{(Left)} Existing benchmarks evaluate TTA methods using streams of images depicting different objects across batches, with random corruptions applied independently to each image.
    \textit{(Right)} Our proposed \shortdatasetname~benchmark addresses these limitations by (i)~presenting images of the same object in a sequence, preserving temporal dependencies from real-world tracklets, and (ii)~applying consistent corruptions whose intensity may evolve over time. This framework offers a more realistic setting for evaluating the adaptability of TTA methods.
  }
    \label{fig:pullfig}
\end{figure*}

\noindent{\textbf{Test-Time Adaptation and its Challenges.}}
Test-Time Adaptation (TTA) seeks to mitigate performance degradation by adapting a pre-trained model on-the-fly using an incoming data stream~\cite{liang2023comprehensive}. At inference, TTA methods perform online unsupervised learning to adjust model parameters in response to new data~\cite{DBLP:journals/corr/abs-2002-08546, sun2020test, wang2020tent, iwasawa2021test}. While TTA has shown promise, current benchmarks oversimplify the problem, primarily simulating distribution shifts without accounting for temporal dependencies that violate the \iid~assumption. 
For instance, many TTA approaches~\cite{DBLP:journals/corr/abs-2002-08546} assume that distribution shifts are purely covariate shifts~\cite{candela2009dataset}, as seen in datasets like CIFAR10-C and ImageNet-C~\cite{imagenetc} (Fig.~\ref{fig:pullfig}, left). Meanwhile, other methods~\cite{Boudiaf_2022_CVPR} address non-\iid~scenarios by modifying label distributions while neglecting the visual continuity inherent in sequential data. A more recent effort by Yuan~\etal~\cite{yuan2023robust} considers both distribution shifts and non-\iid~labels, but still overlooks the critical role of temporal dependencies.
We argue that the lack of benchmarks that jointly capture distribution shifts and temporal dependencies has limited the development of deployable TTA methods. To bridge this gap, we take inspiration from object tracking to introduce a benchmark that inherently accounts for both challenges.

\noindent{\textbf{Introducing \shortdatasetname.}}
Our benchmark, \datasetname~(\shortdatasetname), is built using tracklets—short sequences of images tracking the same object across consecutive frames. By leveraging TrackingNet~\cite{muller2018trackingnet}, we create a realistic test-time adaptation setting where temporal dependencies naturally emerge, leading to \iid~violations. To introduce controlled distribution shifts, we apply standard transformations and corruptions~\cite{imagenetc,3dcc} (\eg Gaussian noise, glass blur) consistently across tracklets rather than as independent perturbations (Fig.~\ref{fig:pullfig}, right). 
This setup better reflects real-world challenges by aligning the temporal structure of the dataset with the sequential nature of data streams, following a protocol inspired by RoTTA~\cite{yuan2023robust}. Using \shortdatasetname, we rigorously analyze how temporal dependencies interact with distribution shifts, revealing significant weaknesses in existing TTA methods.

\noindent{\textbf{Advancing TTA with \shortadvmeminitname.}}
Our investigation highlights that most TTA methods struggle under the compounded effects of distribution shifts and temporal dependencies, leading to severe performance drops. We examine memory-bank-based methods, which should, in principle, handle temporal challenges well due to their ability to adapt to selectively stored samples. However, our results show that these methods suffer from poor initialization of the memory bank, significantly impacting performance.

To address this, we propose \shortadvmeminitname, a novel adversarial memory initialization strategy that enhances stability in adaptation. By leveraging synthetic noise generated in a class-diverse manner, \shortadvmeminitname~operates as a plug-and-play enhancement for memory-based TTA methods. Experiments demonstrate its effectiveness—equipping SHOT-IM \cite{shot}, with \shortadvmeminitname~reduces error rates by 44\% in Tracklet-Wise \iid~settings (see Table~\ref{tab:tracklet_iid_adv_mem}).
\newline

\noindent{\textbf{Our Contributions.}}
\begin{itemize}[-]
    \item \textbf{\shortdatasetname~Benchmark.} We introduce \datasetname, a novel benchmark for TTA that integrates object tracklets, capturing real-world distribution shifts and temporal dependencies.
    \item \textbf{Comprehensive Evaluation of TTA Methods.} We systematically assess existing TTA approaches on \shortdatasetname, highlighting their limitations under realistic non-\iid~conditions.
        \item \textbf{\shortadvmeminitname.}  We equip existing TTA methods with memory and benchmark their memory-adapted versions, demonstrating the impact of incorporating memory mechanisms on adaptation performance. Additionally, we propose an adversarial memory initialization strategy that significantly improves the performance, particularly under severe non-\iid~scenarios.

\end{itemize}

\noindent Our work advances the study of test-time adaptation by providing a more realistic evaluation framework and a novel solution to enhance the stability of model adaptation in dynamic environments.

\begin{figure*}[ht]
    \centering
    \vspace{-0.65cm}
    \includegraphics[width=0.8\textwidth]
    {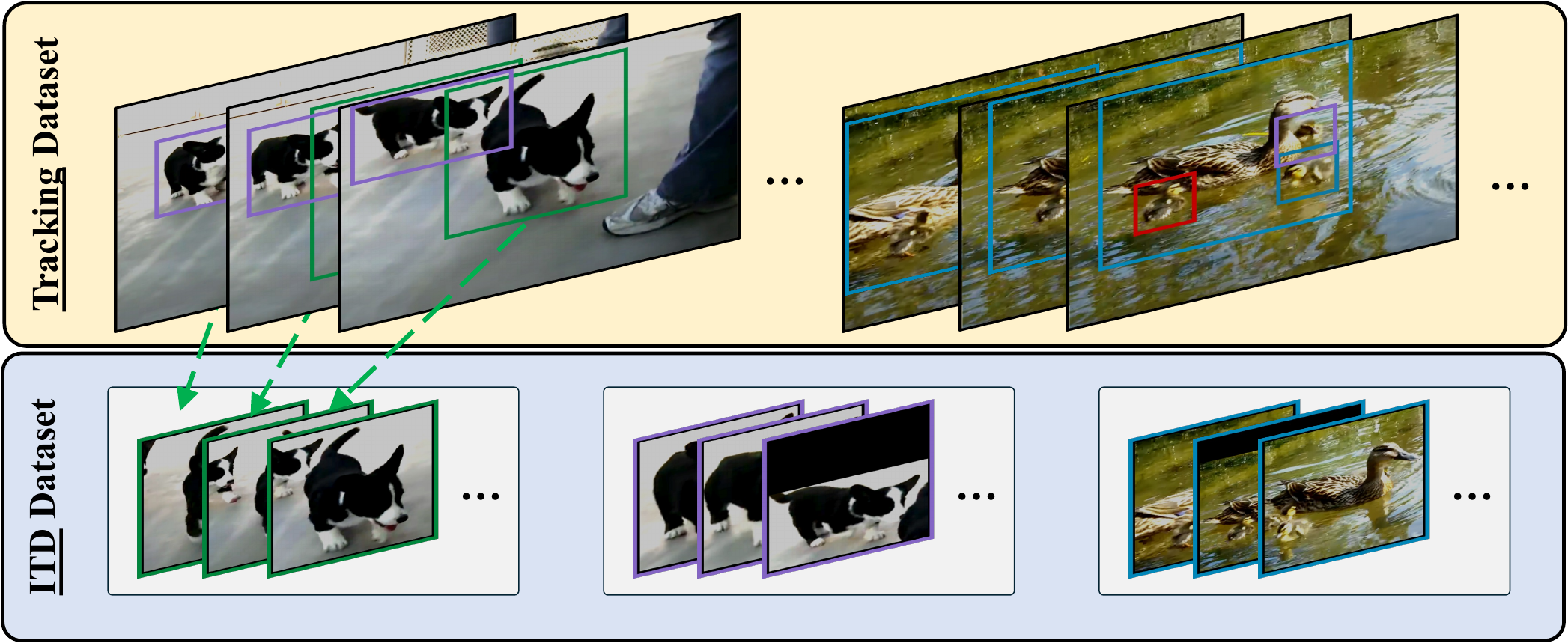}
    \caption{
    \textbf{We build \shortdatasetname~with realistic TTA instances by constructing them from a tracking dataset.}
    We extract object-centric sequential video frames, encapsulating the small variations of the same entity over time. 
    We source the frames and bounding boxes from TrackingNet, a well-established tracking dataset, such that the instances focus on particular objects of interest.
    As such, these sequences naturally exhibit the temporal dependencies inherent in real-world scenarios.
    \label{fig:tracklets}}
    \vspace{-0.5cm}
\end{figure*}
\section{Related Work}
\label{sec:related_work}
\noindent{\textbf{Test-Time Adaptation.}}
TTA leverages the unlabeled data that arrives at test time
to adapt the forward pass of pre-trained DNNs according to some proxy task~\cite{tent, shot}.
Many existing TTA methods focus on 
covariate distribution 
shifts~\cite{adabn, sar, eata, cotta, shot}. 
Several TTA methods tackle this challenge by updating the statistics of the Batch Normalization layers at test time~\cite{adabn, bnadaptation}. 
For example, AdaBN~\cite{adabn} introduces Adaptive Batch Normalization, an algorithm to adapt to the target domain. 
Another group of methods uses an entropy minimization strategy. 
For instance, TENT~\cite{tent} minimizes the entropy of the model's predictions. 
ETA and EATA~\cite{eata} extend TENT by selecting reliable and non-redundant samples to update the model weights. 
More recently, RoTTA~\cite{yuan2023robust} attempts to combat non-\iid~streams at test time by leveraging a memory bank for adapting to an incoming stream of data.
In this work, we introduce \shortadvmeminitname, a novel adversarial memory initialization strategy to significantly enhance the adaptability of TTA methods under complex, non-\iid~scenarios.

\noindent{\textbf{Benchmarking TTA Methods.}}
The fundamental premise of TTA involves deploying a pre-trained model 
onto edge devices like self-driving cars or surveillance cameras, where it faces potential changes in data distribution~\cite{DBLP:journals/corr/abs-2002-08546,ttt,wang2020tent,iwasawa2021test}. 
This scenario unfolds as the model 
encounters a continuous stream of data, with each input potentially coming from a distribution different than the one 
the model was originally trained on.
To emulate such scenarios, the TTA literature commonly creates a stream of data with samples from the test set of well-established image classification datasets, such as ImageNet~\cite{deng2009imagenet} and CIFAR~\cite{krizhevsky2009learning}.
Setups then systematically simulate covariate distribution shifts by inducing corruptions on individual images, such as those from Common Corruptions \cite{imagenetc} and 3D Common Corruptions \cite{3dcc}.
In this work, we present a comprehensive benchmark for simulating more realistic and complex scenarios.

\section{Dataset and Methodology}
\label{sec:methodology}

\noindent{\textbf{Motivation.}} Recent advancements in the field of Test Time Adaptation~(TTA) have departed from traditional independent and identically distributed (\iid) setups towards more nuanced non-\iid~configurations. RoTTA \cite{yuan2023robust} introduced correlation sampling to enforce non-\iid~distributions in labels, motivated by real-world deployment scenarios—e.g., in edge devices, where objects in a scene often appear with correlated labels (e.g., frequent occurrences of ``pedestrian" in a crowded area). This shift revealed a critical limitation: existing TTA methods struggle when faced with such non-\iid~streams.
\begin{table}[]
\centering
\caption{\textbf{Average Error Rates on CIFAR-10-C:} Comparison between non-\iid~episodic evaluation and tracklet mimic evaluation averaged across corruptions. The tracklet mimic setting simulates real-world temporal dependencies. }
\label{tab:tta_cifar10c_comparison}
    \scriptsize
\resizebox{\columnwidth}{!}{%

\begin{tabular}{lccc}
\toprule
TTA Method & non~\iid (\%) & Tracklet Mimic (\%) & $\Delta$ (\%) \\
\midrule

    Source & 44.1 &\textbf{44.1} & \textbf{0}\\
    \midrule
    AdaBN &                 75.4 &                                78.7 &  3.3 \\
        CoTTA &                 75.5 &                                89.1 &             13.6 \\
          SAR &                 75.2 &                                82.4 &              7.2 \\
          ETA &                 75.4 &                    78.6 &     \underline{3.2} \\
         TENT &                 75.3 &                                85.1 &              9.8 \\
        RoTTA &     \underline{27.6} &                       \underline{66.8} &             39.2 \\
\bottomrule
\end{tabular}%
}
\end{table}

In real-world streaming data (e.g., videos from surveillance cameras), consecutive frames often depict the same object with minor variations, leading to visual redundancy. To highlight the impact of temporal dependencies on TTA performance, we conducted a simple experiment on CIFAR-10-C. In this experiment, we compared the error rates of several TTA methods under PTTA \cite{yuan2023robust} evaluation against a modified setup where each batch contains images duplicated to mimic a video clip or tracklet. As shown in Table~\ref{tab:tta_cifar10c_comparison}, all methods experience a noticeable performance drop in the tracklet mimic setting. This finding clearly illustrates the challenges posed by temporal dependencies and motivates the need for our new \shortdatasetname~benchmark, which better reflects real-world data streams.

These observations highlight the need for adaptation strategies robust to \textit{visual} non-\iid~shifts and a benchmark that captures these complexities. To this end, we introduce ITD, a benchmark built explicitly for evaluating methods under complex non-\iid~scenarios, and propose \shortadvmeminitname, a plugin designed to mitigate over-adaptation in such conditions.

\noindent\textbf{Tracklets: A Natural Source of Visual Non-\iid Data.} 
To construct a realistic non-\iid~benchmark, we leverage the field of Object Tracking. Unlike artificially simulated correlation sampling used in prior works \cite{yuan2023robust}, object-tracking datasets inherently capture realistic temporal dependencies by tracking specific objects across video frames. We propose using \textit{tracklets}—sequences of object-centric images extracted from tracking datasets to model the gradual variations encountered in real-world image streams. This approach not only enhances the realism of our benchmark but also faithfully captures intrinsic characteristics of natural image sequences, establishing a robust foundation for evaluating TTA in real-world scenarios.
\subsection{Dataset Construction}
We construct \datasetname~(\shortdatasetname) from a large-scale object-tracking dataset. Specifically, we utilize TrackingNet~\cite{muller2018trackingnet}, which originates from the YouTube Bounding-Boxes dataset~\cite{DBLP:journals/corr/RealSMPV17}. 

\noindent For each video, we extract tracklets by iterating through frames where a given object appears, cropping bounding boxes around it. This process results in \shortdatasetname, a dataset composed of object tracklets—sequences of images capturing realistic temporal variations of objects. Figure~\ref{fig:tracklets} illustrates the tracklet extraction process. 

\noindent In practice, we apply the following preprocessing steps:
\begin{itemize}[-]
    \item \textbf{Frame Selection:} Extract crops at a 5-frame interval to balance dataset size and temporal redundancy.
    \item \textbf{Crop Resizing:} Extract square crops 10\% larger than the largest side of the bounding box to retain contextual information.
    \item \textbf{Standardization:} Resize all crops to 224$\times$224 for consistency and compatibility with batch-based training.
\end{itemize}
These design choices ensure a realistic yet tractable dataset, allowing for systematic evaluation of TTA methods.

\noindent\textbf{Dataset Properties.} Unlike conventional datasets where samples are independent, \shortdatasetname~is composed of \textit{tracklets}, preserving the temporal continuity of objects. Each tracklet consists of images depicting the same object in different frames, naturally encoding non-\iid~dependencies. Additionally, our dataset supports temporally consistent corruptions (detailed in Section~\ref{sec:prep_tta}), further enhancing its relevance for evaluating TTA under realistic conditions.

\noindent\textbf{Statistics.} The \shortdatasetname~dataset contains over 23$\texttt{K}$ objects that span 21 classes. The dataset is divided into training (50\%), validation (30\%), and test (20\%) sets. In total, it comprises over 220$\texttt{K}$ images—more than four times the size of ImageNet-C~(50$\texttt{K}$)—while also providing object-instance relationships via tracklets. Further statistics, including class distributions, are provided in the appendix underscoring the scale and diversity of \shortdatasetname, making it a valuable resource for advancing TTA research.

\section{Benchmarking on \shortdatasetname}\label{sec:benchmark}
\noindent\textbf{Overview.} Unlike previous benchmarks that assume independent samples, \shortdatasetname introduces a tracklet-based evaluation to reflect real-world challenges, such as sequential dependencies and non-\iid~distributions. We systematically evaluate TTA methods across three levels of complexity to assess their adaptability in streaming environments:

\begin{figure*}[h]
    \centering
    \includegraphics[width=\textwidth]{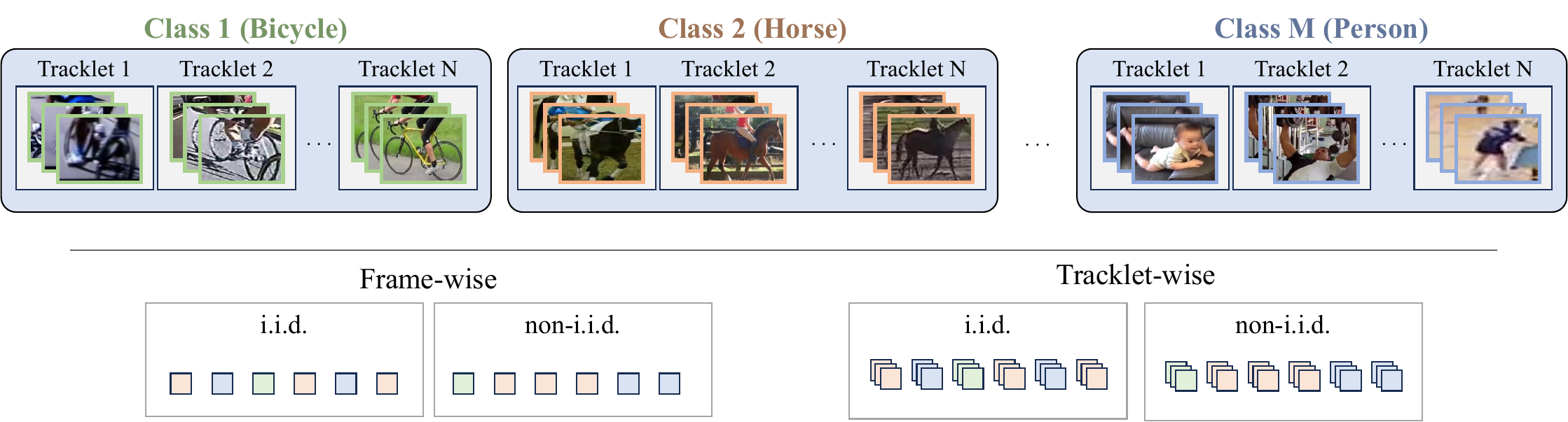}
    \caption{\textbf{Frame-wise and Tracklet-wise Experiment Setup:} We illustrate the construction of the frame-wise and tracklet-wise experiments. In the frame-wise setup, one frame is sampled from each tracklet to ensure each object is observed once. In the tracklet-wise setup, the frames within each tracklet are sequentially processed. Both \iid and non-\iid settings are depicted for each setup.}
    \label{fig:setups}

\end{figure*}
\begin{itemize}
    \item \textbf{Frame-wise \iid~(Section~\ref{sec:frams_iid}):} Frames are sampled independently and identically distributed (\iid), without considering sequential dependencies.
    \item \textbf{Tracklet-wise \iid~(Section~\ref{sec:tracklet_iid}):} Entire tracklets are sampled \iid, preserving intra-tracklet dependencies while maintaining inter-tracklet randomness.
    \item \textbf{Tracklet-wise non-\iid~(Section~\ref{sec:tracklet_non_iid}):} Tracklets are sampled following a Dirichlet distribution~\cite{yuan2023robust}, enforcing stronger non-\iid~properties across the dataset.
\end{itemize}

\noindent These setups enable a systematic evaluation of TTA methods under increasingly complex real-world conditions. Each scenario is tested under a single domain shift (e.g., fog) to isolate the effect of adaptation techniques.
\subsection{Tracklet-Based Adaptation Strategies}
Unlike frame-wise adaptation, tracklet-based setups introduce sequential dependencies, making naive entropy minimization unreliable due to biased batch statistics. To ensure fair comparisons, we extend TENT and SHOT-IM by incorporating RoTTA’s memory bank \(\mathcal{M}\). This allows them to store and utilize previously seen samples, ensuring more stable adaptation in non-\iid~streams. In these adaptations, we replace RoTTA’s original objective with entropy minimization for TENT and information maximization for SHOT-IM.

\subsection{Experimental Setup}
Having established the dataset structure, corruption strategies, and adaptation mechanisms, we now outline our standardized experimental setup for evaluating TTA methods. Unless stated otherwise, we use ResNet-18 as the base model, apply corruptions at the highest severity level (5), and set a batch size of 64 for streaming data. We evaluate eight TTA methods and compare them against a pre-trained model 
\(f_{\theta}\)(Source) as the baseline. The details of these methods are provided in Table~\ref{tab:tta_methods}. Each method is assessed using its optimal hyperparameters, determined through an extensive search.

\begin{table}[h]
    \centering
    \caption{Overview of TTA Methods Evaluated in \shortdatasetname.}
    \resizebox{\columnwidth}{!}{%
    \begin{tabular}{l|l}
        \toprule
        \textbf{Method} & \textbf{Adaptation Strategy} \\
        \midrule
        \textbf{AdaBN} \cite{adabn} & Updates batch normalization statistics. \\
        \textbf{SHOT-IM} \cite{shot} & Maximizes mutual information. \\
        \textbf{TENT} \cite{tent} & Utilizes entropy minimization for adaptation. \\
        \textbf{SAR} \cite{sar} & Employs sharpness-aware optimization. \\
        \textbf{EATA} \cite{eata} & Entropy-based sample selection. \\
        \textbf{CoTTA} \cite{cotta} & Uses consistency-based distillation for continual adaptation. \\
        \textbf{RoTTA} \cite{yuan2023robust} & Maintains a memory bank to stabilize adaptation. \\
        \bottomrule
    \end{tabular}}
    \label{tab:tta_methods}
\end{table}

\subsection{Considered Corruptions}
\begin{table}[!h]
    \centering
    \caption{\textbf{Types of Corruptions Applied in \shortdatasetname.} To evaluate TTA robustness, we introduce common distribution shifts from ImageNet-C.}
    \resizebox{\columnwidth}{!}{%
    \begin{tabular}{l l}
        \toprule
        \textbf{Category} & \textbf{Corruptions} \\
        \midrule
        \textbf{Noise} & Gaussian, Shot, Impulse \\
        \textbf{Blur} & Defocus, Glass, Zoom \\
        \textbf{Weather} & Snow, Frost, Fog, Brightness \\
        \textbf{Digital Artifacts} & Contrast, Elastic Transform, Pixelate, JPEG Compression \\
        \bottomrule
    \end{tabular}%
    }
    \label{tab:corruptions}
\end{table}

Corruptions, detailed in Table~\ref{tab:corruptions}, are applied dynamically as the data stream unfolds. Given the temporal nature of \shortdatasetname, we also explore scenarios where corruption severity varies within a tracklet, simulating transient environmental fluctuations (e.g., changing weather conditions). Additional extended evaluations can be found in the appendix.

\subsection{Preparation for TTA}\label{sec:prep_tta}
To evaluate the effectiveness of TTA methods on our \shortdatasetname~benchmark, we require models that are either pre-trained or fine-tuned specifically on this dataset. While \shortdatasetname~shares some class overlap with ImageNet, its distribution differs significantly, limiting the effectiveness of direct transfer. Our experiments confirm that ImageNet-pretrained models struggle to generalize well to \shortdatasetname, highlighting the need for dataset-specific fine-tuning. Therefore, we fine-tune two ImageNet-pretrained models, ResNet-18 and ViT-B-16, on \shortdatasetname's training set (see Table~\ref{tab:model_comparison}).

\begin{table}[h]
    \centering
    \caption{\textbf{Error rates on all splits of \shortdatasetname.} Detailed training and test loss/accuracy results are provided in the appendix.}
    \scriptsize
    \renewcommand{\arraystretch}{1.2} 
    \setlength{\tabcolsep}{12pt} 
    \begin{tabular}{l|cc}
        \toprule 
        \textbf{Error Rate} & \textbf{ResNet-18} & \textbf{ViT}  \\
        \midrule
        Train & 4.1 & 2.0  \\
        Validation & 8.2 & 3.8  \\
        Test & 9.4 & 4.0  \\
        \bottomrule
    \end{tabular}
    \label{tab:model_comparison}
\end{table}
\subsection{Frame-Wise \iid~Scenario}\label{sec:frams_iid}
\begin{table*}[t]
\centering
\caption{\textbf{Performance Comparison under Frame-wise \iid~Assumption.} 
Average error rates reported for TTA methods on images from the \textit{\shortdatasetname}~dataset. 
In this setup, all images/frames are shuffled and then independently subjected to corruptions. 
Notably, SHOT-IM outperforms other methods across all corruptions, showcasing its robustness against these domain shifts.}
\centering
\scriptsize
\begin{tabularx}{\textwidth}{c|XXXXXXXXXX}
\toprule
M\textbf{ethod}&\textbf{Source} & \textbf{AdaBN} & \textbf{SHOT}-IM & \textbf{TENT} & \textbf{SAR} & \textbf{CoTTA} & \textbf{ETA} & \textbf{EATA} & \textbf{RoTTA} \\
\midrule
Avg. Err.~\ensuremath{\downarrow}
& 62.4 & 46.9 & \textbf{39.3} & 46.8 & 46.5 & 46.7 & 46.7 & 46.8 & 53.4 \\
\bottomrule\bottomrule
\end{tabularx}
\label{tab:fram_iid}
\end{table*}
In this scenario, we evaluate TTA methods under a conventional setting where each frame is treated as an independent and identically distributed (\iid) sample, ignoring temporal dependencies. This setup assumes a uniform label distribution across time, thereby oversimplifying real-world conditions where label distributions may be highly imbalanced.

\noindent To construct the test stream, we shuffle tracklets and sample one frame per tracklet, eliminating the notion of temporal continuity. Following standard practice \cite{imagenetc, 3dcc}, we assess performance degradation under 15 different corruptions.
Table~\ref{tab:fram_iid} presents the average error rates of the eight evaluated TTA methods. As expected, distribution shifts significantly degrade the performance of pre-trained models. For example, the error rate of the Source model (ResNet-18 without TTA) rises from below 10\% (Table~\ref{tab:model_comparison}) to over 60\% (Table~\ref{tab:fram_iid}). Notably, TENT \cite{tent} reduces the error rate to below 50\% through entropy minimization, while SHOT-IM achieves the lowest error, averaging below 40\%.

\subsection{Tracklet-Wise \iid~Scenario}\label{sec:tracklet_iid}
To move towards a more realistic evaluation, we introduce a setup where the model processes entire tracklets at test time. Within each tracklet, consecutive frames share labels and contextual consistency, though tracklets are sampled in an \iid~manner. This setup emulates real-world scenarios where the model observes a single object over multiple frames before switching to a new one.

\noindent We evaluate SHOT-IM (the best performer from Section~\ref{sec:frams_iid}), TENT, and RoTTA. Although RoTTA performed poorly in the frame-wise \iid~scenario, its memory-based approach is specifically designed for non-\iid~streams, making it relevant for this setting. To ensure a fair comparison, we extend TENT and SHOT-IM by incorporating RoTTA’s memory bank, allowing them to only adapt to informative samples selected and retained in memory based on RoTTA Category-balanced sampling heuristics.

\noindent Table~\ref{tab:tracklet_iid_mem} summarizes our results. We find that SHOT-IM and TENT struggle under tracklet-based evaluation, with SHOT-IM’s error rate increasing from under 40\% (Table~\ref{tab:fram_iid}) to nearly 95\% (Table~\ref{tab:tracklet_iid_mem}). This decline stems from the biased statistics computed within tracklets, which skew entropy-based adaptation. In contrast, RoTTA demonstrates superior stability, reducing the average error to around 50\%, due to its distillation-based approach.

\begin{table}[ht]

\small
\centering
\caption{\textbf{Tracklet-wise \iid~and Tracklet-wise \textit{non}-\iid~Evaluation with and without Memory.} 
We report average error rates for TTA methods on our \shortdatasetname~dataset under both the tracklet-wise \iid~scenario (\ie entire tracklets are sampled \iid) and when tracklets are non-\iid, \ie tracklets are sampled such that their labels display correlation in time~(by following a Dirichlet distribution). 
The results are grouped to reflect the presence (\cmark) or absence (\xmark) of memory during adaptation. 
RoTTA shows notable robustness by using memory, significantly outperforming the non-memory variants across various corruption types. Under \textit{non-}\iid. RoTTA showcases it's proficiency against non-\iid~data streams.}
\resizebox{\columnwidth}{!}{

\begin{tabular}{l|c|cc}

\toprule
 &&Error Rate\ensuremath{\downarrow} 
&Error Rate \ensuremath{\downarrow} 
\\
\multirow{-2}{*}{Method} &\multirow{-2}{*}{Memory}  &Tracklet-wise \iid~& Tracklet-wise \textit{non}-\iid\\

\midrule
TENT & \xmark &         94.0 & 94.0\\
      & \cmark &             93.8 & 93.8\\\midrule
SHOT-IM & \xmark &     94.7&  95.1 \\
      & \cmark &  93.4&  93.6 \\\midrule
RoTTA & \cmark &\textbf{51.3} & \textbf{79.3} \\
\bottomrule\bottomrule
\end{tabular}
}
\label{tab:tracklet_non_iid_mem}\label{tab:tracklet_iid_mem}
\vspace{-.5cm}
\end{table}

\subsection{Tracklet-Wise non-\iid~Scenario}\label{sec:tracklet_non_iid}
In this final and most challenging setup, tracklets are sampled non-\iid~to simulate real-world streaming conditions such as autonomous driving or surveillance, where object categories appear in bursts. To model this, we follow Yuan~\etal~\cite{yuan2023robust} and sample tracklets using a Dirichlet distribution \(\textrm{Dir}(\gamma)\). As \(\gamma \rightarrow 0\), label correlation within the stream increases, deviating from the \iid~assumption.
\newline
We evaluate RoTTA, SHOT-IM, and TENT under \(\gamma = 10^{-4}\) to enforce strong non-\iid~conditions (additional \(\gamma\) values are analyzed in Section~\ref{sec:adv_mem}). Table~\ref{tab:tracklet_non_iid_mem} reports the results. Even RoTTA, designed for non-\iid~streams, experiences a 28\% performance drop compared to the tracklet-wise \iid~setup (Table~\ref{tab:tracklet_iid_mem}). We hypothesize that this decline results from imbalanced memory due to empty memory initialization, where certain classes are observed late in the stream, causing forgetting effects.
These findings suggest that memory-based TTA methods can be further improved by introducing class-balancing initialization mechanisms to stabilize adaptation over long, non-\iid~streams

\section{Experiments:\newline Enhancing Performance with Memory}\label{sec:adv_mem}

In Section~\ref{sec:tracklet_iid}, we observed that equipping TENT and SHOT-IM with a memory bank, while seemingly beneficial, does not yield significant performance improvements. Furthermore, in Section~\ref{sec:tracklet_non_iid}, we demonstrated how a tracklet-wise non-\iid~stream severely impacts RoTTA's performance. These findings indicate that while memory banks are essential for adapting to non-\iid~streams~\cite{yuan2023robust}, even strong TTA methods experience significant degradation when evaluated on \shortdatasetname.

\noindent We hypothesize that this degradation stems from the empty initialization of the memory bank. Consider the extreme case where $\gamma \rightarrow 0$, resulting in the memory bank lacking any examples for labels revealed later in the stream until those labels actually appear. Additionally, methods such as TENT rely on statistical measures like entropy for updates. When key classes are absent from the memory bank, model updates become skewed, leading to catastrophic forgetting. This motivates the need for a carefully designed memory initialization strategy that ensures stable adaptation steps.

\subsection{Adversarial Memory Initialization}

To address these challenges, we propose a novel approach for initializing the memory bank to enhance adaptation. Our goal is to populate the memory bank with class-representative samples to:
\textbf{(i)} prevent forgetting by ensuring all classes are accounted for, and \textbf{(ii)} balance computed statistics in the output space.

\noindent To that end, we propose initializing the memory bank with synthetic data generated by adversarial algorithms~\cite{tzeng2017adversarial}. Specifically, each memory bank entry is initialized as Gaussian noise, assigned a random label, and subjected to a targeted adversarial attack that maximizes the network’s confidence in classifying it correctly, following~\cite{alfarra2022combating, goodfellow2014explaining}. Formally, let $\mathcal{M}$ be an initially empty memory bank with a maximum capacity of $N$. We populate $\mathcal{M}$ iteratively with synthetic examples $x^*$, where:
\begin{equation}
    \label{eq:adversarial_attack}
    x^* = \underset{x}{\arg\min} \: \mathcal{L}_{\text{ce}}(f_\theta(x), y),
\end{equation}

\setcounter{table}{1} 

\begin{table*}[!ht]
\caption{
\textbf{Effect of Adversarial Memory Initialization on TTA performance in the Tracklet-Wise \iid~scenario.} 
We evaluate TTA methods on \shortdatasetname~in an \iid~tracklet context, contrasting standard (\xmark) and adversarial (\cmark) memory initialization~(\shortadvmeminitname). 
When we equip SHOT-IM with \shortadvmeminitname, it outperforms all other methods, demonstrating its capacity to enhance adaptability.}
\resizebox{\textwidth}{!}{
\begin{tabular}{lc|ccc|ccc|cccc|cccc|c}

\toprule
 & & \multicolumn{3}{c|}{Noise} & \multicolumn{3}{c|}{Blur}                           & \multicolumn{4}{c|}{Weather} & \multicolumn{4}{c|}{Digital} \\
\multirow{-2}{*}{Method} & \multirow{-2}{*}{\shortadvmeminitname}  & gauss. & shot   & impul.  & defoc.& glass & zoom & snow& frost& fog        & brigh.      & contr.      & elast.      & pixel.      & jpeg      & \multirow{-2}{*}{Avg.} \\\midrule
TENT & \xmark &           94.2 &           94.2 &           94.3 &           94.5 &           94.5 &           93.9 &           93.7 &           93.7 &           93.2 &           93.0 &           92.8 &           93.8 &           93.4 &           93.5 &           93.8 \\
      & \cmark &           85.1 &           83.2 &           88.2 &           82.9 &           84.8 &           74.0 &           85.5 &           84.2 &           83.9 &           68.9 &           71.6 &           79.9 &           62.6 &           69.5 &           78.9 \\\midrule
SHOT-IM & \xmark &           93.8 &           93.9 &           93.9 &           94.2 &           94.3 &           93.7 &           93.3 &           93.0 &           93.0 &           92.8 &           92.1 &           93.5 &           93.4 &           93.2 &           93.4 \\
      & \cmark &  \textbf{61.7} &  \textbf{58.0} &  \textbf{62.3} &  \textbf{52.5} &  \textbf{51.0} &  \textbf{39.7} &           57.2 &  \textbf{60.6} &  \textbf{51.4} &           32.8 &  \textbf{55.0} &  \textbf{38.6} &  \textbf{27.6} &           35.6 &  \textbf{48.9} \\\midrule
RoTTA & \xmark &           68.0 &           62.4 &           68.6 &           58.5 &           56.7 &           40.9 &  \textbf{56.6} &           60.7 &           52.2 &           30.2 &           60.7 &           39.7 &           27.8 &  \textbf{35.1} &           51.3 \\
      & \cmark &           69.0 &           62.8 &           68.9 &           58.1 &           58.0 &           40.9 &           57.2 &           61.0 &           53.0 &  \textbf{30.0} &           62.4 &           41.2 &           27.7 &           35.8 &           51.9 \\
\bottomrule
\bottomrule
\end{tabular}
}
\label{tab:tracklet_iid_adv_mem}
\end{table*}

\begin{table*}[!ht]

\caption{
\textbf{Effect of Adversarial Memory Initialization on TTA performance in the Tracklet-Wise \textit{non}-\iid~scenario.} 
We evaluate methods on \shortdatasetname~in a non-\iid~tracklet setting, comparing standard (\xmark) and adversarial (\cmark) memory initialization~(\shortadvmeminitname). 
RoTTA with \shortadvmeminitname~significantly outperforms the alternatives. 
}
\resizebox{\textwidth}{!}{
\begin{tabular}{lc|ccc|ccc|cccc|cccc|c}
\toprule
 &  & \multicolumn{3}{c|}{Noise} & \multicolumn{3}{c|}{Blur}                           & \multicolumn{4}{c|}{Weather} & \multicolumn{4}{c|}{Digital} \\
\multirow{-2}{*}{Method} & \multirow{-2}{*}{\shortadvmeminitname}  & gauss. & shot   & impul.  & defoc.& glass & zoom & snow& frost& fog        & brigh.      & contr.      & elast.      & pixel.      & jpeg      & \multirow{-2}{*}{Avg.} \\
\midrule
TENT & \xmark &           94.2 &           94.3 &           94.3 &           94.5 &           94.5 &           94.0 &           93.7 &           93.6 &           93.4 &           93.0 &           92.9 &           93.8 &           93.4 &           93.6 &           93.8 \\
      & \cmark &           89.2 &           89.8 &           92.4 &           92.1 &           92.4 &           86.6 &           91.4 &           90.2 &           90.7 &           83.7 &           83.9 &           89.0 &           79.5 &           81.5 &           88.0 \\\midrule
SHOT-IM & \xmark &           93.9 &           93.8 &           93.9 &           94.3 &           94.3 &           93.7 &           93.4 &           93.6 &           93.2 &           92.7 &           92.5 &           93.6 &           93.5 &           93.2 &           93.6 \\
      & \cmark &           92.3 &           92.2 &           92.7 &           92.4 &           93.1 &           91.6 &           92.5 &           91.8 &           91.5 &           90.5 &           92.0 &           91.6 &           89.1 &           90.9 &           91.7 \\\midrule
RoTTA & \xmark &           85.6 &  \textbf{81.6} &           86.5 &           86.9 &           87.3 &           78.9 &           81.7 &           83.5 &           75.0 &           67.2 &  \textbf{71.8} &           79.7 &           77.1 &           67.4 &           79.3 \\
      & \cmark &  \textbf{84.4} &           82.2 &  \textbf{84.1} &  \textbf{83.0} &  \textbf{83.3} &  \textbf{68.6} &  \textbf{80.2} &  \textbf{80.5} &  \textbf{74.7} &  \textbf{62.8} &           73.6 &  \textbf{75.2} &  \textbf{61.3} &  \textbf{62.9} &  \textbf{75.5} \\
\bottomrule

\bottomrule
\end{tabular}
}
\label{tab:tracklet_non_iid_adv_mem}

\end{table*}

\noindent where $y$ is a randomly assigned label, and $\mathcal{L}_{\text{ce}}$ represents the cross-entropy loss. We solve this optimization problem by applying gradient descent, starting from Gaussian noise. This process is repeated $N$ times to fully initialize $\mathcal{M}$. We term this procedure ``\shortadvmeminitname'' and present it in Algorithm~\ref{alg:memory_init}.
\noindent Our adversarial memory initialization offers two key advantages:
\textbf{(i)} The memory bank remains populated with class-representative samples throughout adaptation, mitigating forgetting under strong non-\iid~streams. 
\textbf{(ii)} The initialization method is independent of how $\mathcal{M}$ is updated or utilized. For example, when applied to RoTTA, the adaptation and update mechanisms remain unchanged.
A straightforward alternative would be initializing $\mathcal{M}$ with uniformly sampled, non-corrupted training examples. However, even in cases where privacy concerns are not an issue, our experiments (detailed in the appendix) indicate that this approach does not improve performance.

\noindent\textbf{Performance with \shortadvmeminitname.}
We implement \shortadvmeminitname initialization strategy within RoTTA, TENT, and SHOT-IM, replacing their default empty memory initialization with adversarially generated samples. We then evaluate on the experimental setups from Sections~\ref{sec:tracklet_iid} and~\ref{sec:tracklet_non_iid} to assess its impact on adaptation performance.
\begin{algorithm}[]
\caption{\shortadvmeminitname}
\label{alg:memory_init}
\begin{algorithmic}
\Function{InitializeMemory}{$f_{\theta}, K, N$}
\State Initialize $\mathcal M = \{\}$
\While{$|\mathcal M| < N$}
\State $x \sim \mathcal N(0, I)$, $y\sim \mathcal U\{1, 2, \dots, K\}$
\While{$f_{\theta}(x) \neq y$}
\State $x \gets x - \alpha \cdot \left(\nabla_x \mathcal{L}_{\text{ce}}(f_{\theta}(x), y)\right)$
\EndWhile
\State $\mathcal M \gets \mathcal M \cup x$
\EndWhile
\State \textbf{return} $\mathcal M$
\EndFunction
\end{algorithmic}
\end{algorithm}
\noindent\textbf{Tracklet-wise \iid~Setup:} 
We first analyze the Tracklet-wise \iid~setting from Section~\ref{sec:tracklet_iid} (i.e., $\gamma \rightarrow \infty$). Table~\ref{tab:tracklet_iid_adv_mem} reports the results, showing significant performance improvements across all baselines. Notably, \shortadvmeminitname~reduces the average error rate of TENT by \(\sim\)15\% and that of SHOT-IM by over 40\%, making SHOT-IM the top-performing method. These improvements highlight the effectiveness of \shortadvmeminitname~in stabilizing adaptation.

\noindent\textbf{Tracklet-wise non-\iid~Setup:} 
To further examine its impact, we evaluate \shortadvmeminitname~in the extreme non-\iid~setting from Section~\ref{sec:tracklet_non_iid} ($\gamma \rightarrow 0$). Table~\ref{tab:tracklet_non_iid_adv_mem} presents the results, demonstrating substantial performance gains. In particular, \shortadvmeminitname~enhances RoTTA’s performance by an average of 4\%, with specific improvements of 16\% and 10\% against pixelate and zoom corruptions, respectively. This trend holds for other methods as well, with TENT improving by over 10\% on JPEG corruption and achieving a 5\% average improvement across all corruptions.

\begin{figure*}[!h]
    \centering
    \subcaptionbox{Transitioning from \textit{non}-\iid~to \iid.\label{fig:gamma_study}}
    {\begin{subfigure}[a]{0.49\textwidth}
        \centering
        \hspace{-1cm}\includegraphics[height=0.4\textwidth]{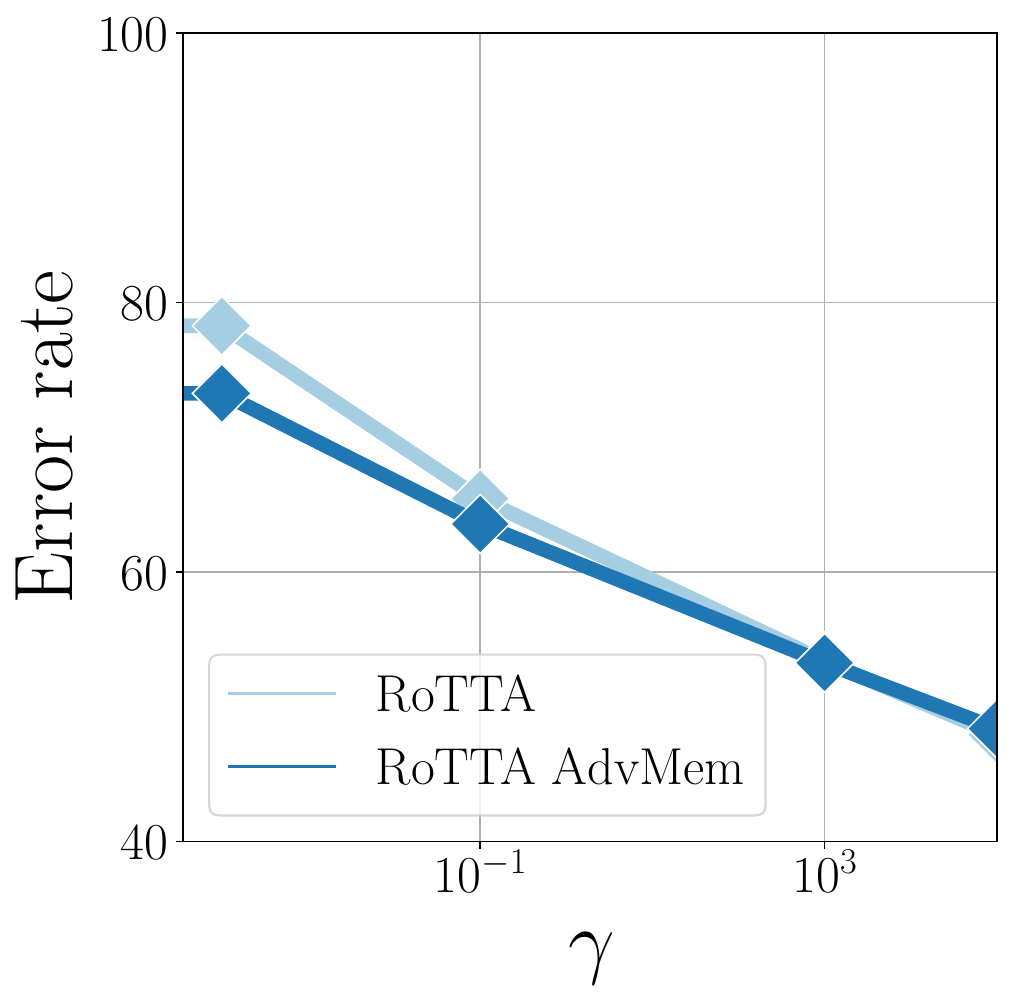}
        \hspace{.25cm}
        \includegraphics[height=0.4\textwidth]{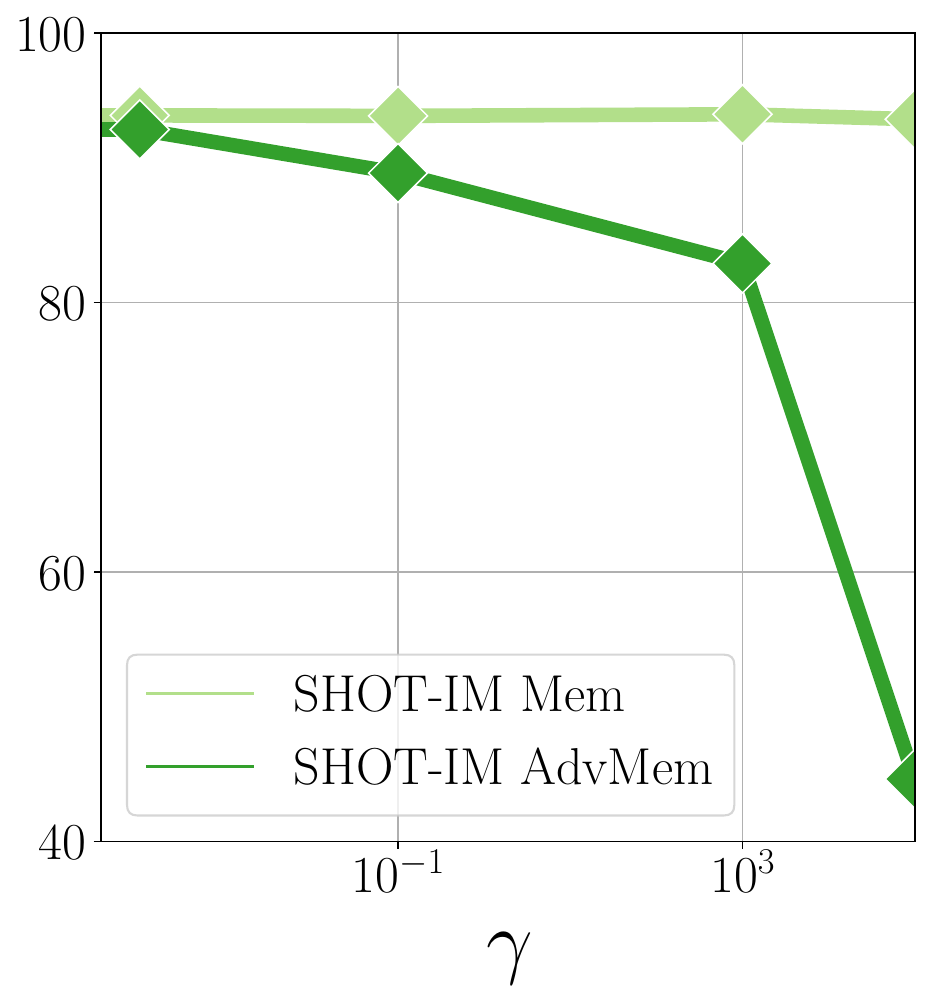}
    \end{subfigure}}  
    \hspace{-2cm}
    \subcaptionbox{Influence of batch size.\label{fig:batch_study}}{\begin{subfigure}[b]{0.49\textwidth}
        \centering
        \includegraphics[height=0.4\textwidth]{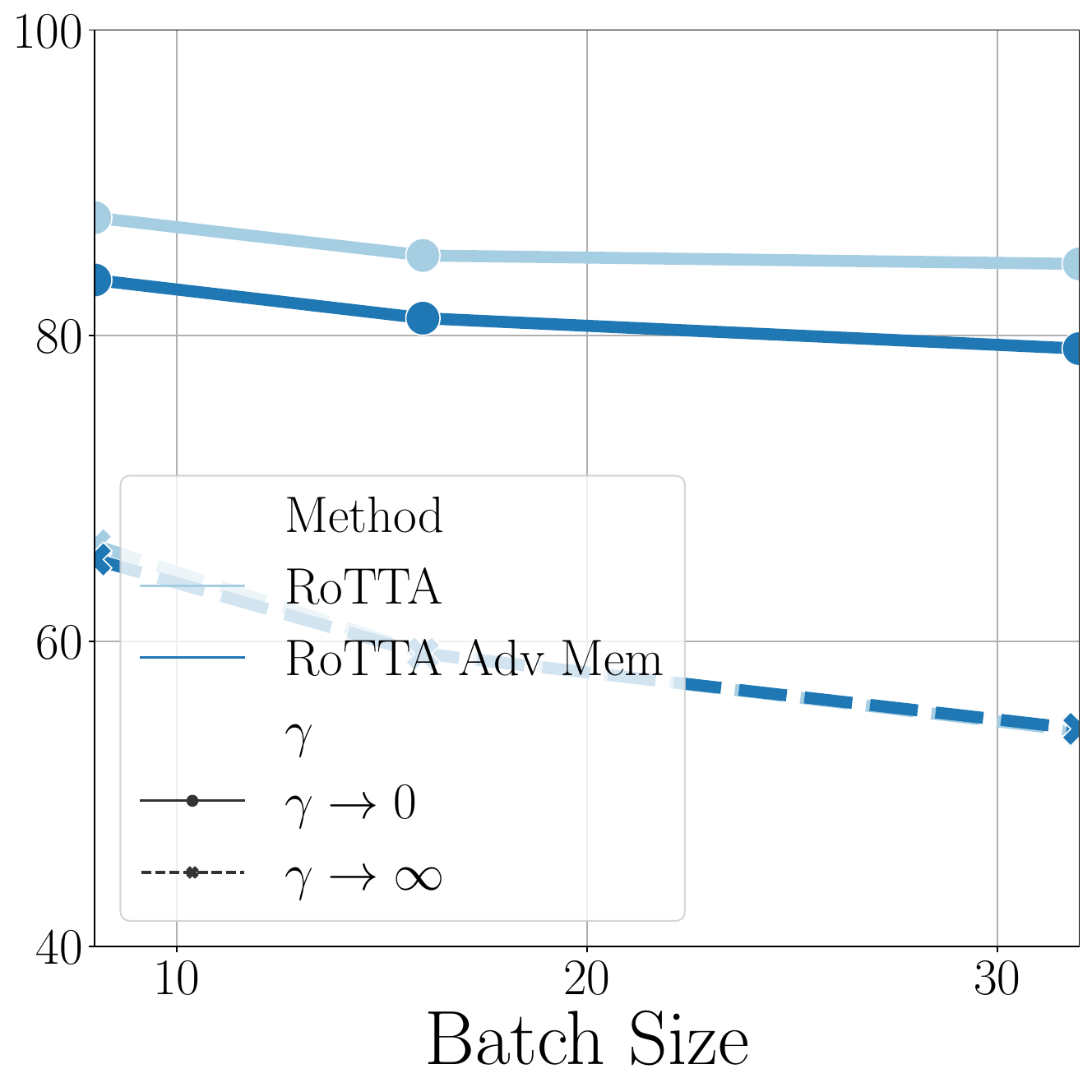}
        \hspace{.25cm}
        \includegraphics[height=0.4\textwidth]{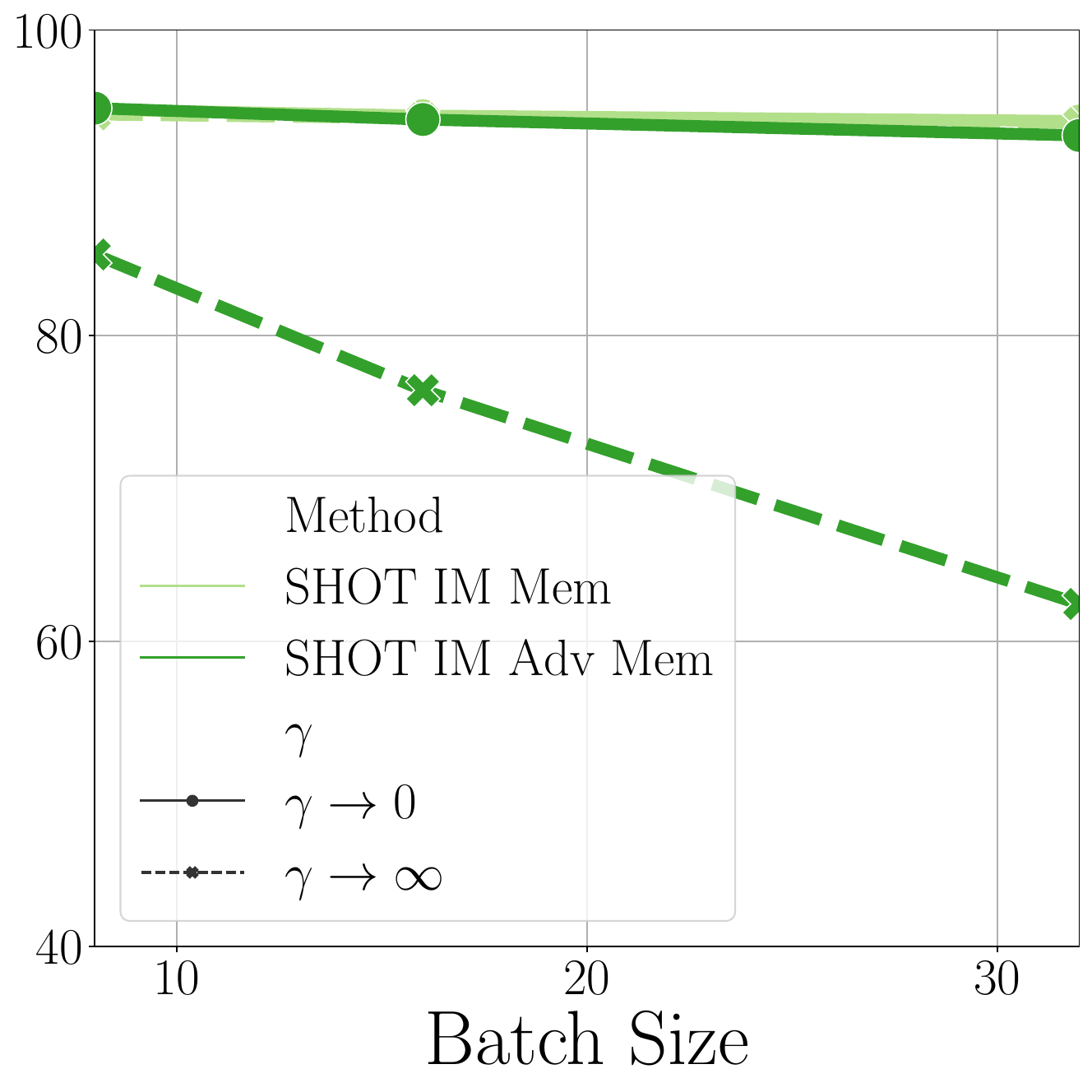}\hspace{-1cm}
    \end{subfigure}}
    \vspace{0.5cm}
\caption{
\textbf{Error rate as \textbf{(a)} we transition the evaluation scenario from non-\iid~to \iid, and \textbf{(b)} as we vary the batch size.}
In \textbf{(a)}, we control the \iid~nature of the label distribution by varying the $\gamma$ parameter in the $\textrm{Dir}(\gamma)$ distribution ($\gamma$-axis in log-scale).
Adding our proposed \shortadvmeminitname~consistently enhances or maintains performance across different \iid-ness regimes and methods.
In \textbf{(b)}, we examine how batch size influences method performance, showing that larger batch sizes improve performance, with \shortadvmeminitname~further enhancing results.
All results presented here are for ResNet-18.
}

\label{fig:adv_plot_gamma}
\end{figure*}
\section{Ablation Studies and Analysis}
\label{sec:exp}
This section presents an in-depth analysis of the \shortdatasetname~dataset and our proposed \shortadvmeminitname. We specifically evaluate both the ResNet-18 and Vision Transformer (ViT) architectures to investigate the impact of \shortadvmeminitname on performance across different experimental conditions.

\subsection{Controlling Label Distribution}\label{sec:label_distrib}

Using a Dirichlet distribution $\textrm{Dir}(\gamma)$ for sampling labels, as introduced in \cite{yuan2023robust}, allows control over the distributional shift in the label space through the parameter $\gamma$. Specifically, as $\gamma \rightarrow 0$, we approach a class-incremental non-\iid~setup, while as $\gamma \rightarrow \infty$, we transition towards a uniform \iid~setup. 

We extend our experiments by analyzing intermediate stages with $\gamma \in \{10^{-4}, 10^{-1}, 10^{3}\}$ and report the results in Figure~\ref{fig:gamma_study}. 

For low values of $\gamma$, the model encounters a challenging class-incremental non-\iid~scenario. In this case, \shortadvmeminitname proves instrumental in mitigating class bias within the incoming image stream, reducing degradation in performance. Conversely, as $\gamma$ increases, the scenario becomes easier, as the model has higher chances of encountering uniformly sampled classes. In this context, the adversarially initialized memory samples are rapidly replaced by reliable examples from the stream, making the initialization inconsequential. Consequently, \shortadvmeminitname neither enhances nor degrades performance in the uniform \iid~setup.

\subsection{Sensitivity to Batch Size}
In this section, we analyze the impact of batch size on performance by varying it across $\{8, 16, 32\}$. The results, reported in Figure~\ref{fig:batch_study}, confirm that increasing the batch size improves performance, as larger batches expose models to more diverse examples, facilitating better adaptation.

When comparing \iid~and non-\iid~setups ($\gamma \rightarrow \infty$ vs. $\gamma \rightarrow 0$), we observe that methods incorporating \shortadvmeminitname~consistently achieve improved or maintained performance across batch sizes. However, as in Section~\ref{sec:label_distrib}, the impact of \shortadvmeminitname~is less pronounced in the \iid~setup, indicating that memory initialization has a limited effect in this scenario.

These insights highlight the robustness of \shortadvmeminitname in highly non-\iid~environments while demonstrating that its advantages diminish in simpler, uniform settings. Overall, the findings reinforce the necessity of well-designed memory initialization strategies in real-world, dynamically shifting data distributions.

\section{Conclusion}
This work introduces \shortdatasetname, a novel benchmark designed to challenge existing TTA methods. Unlike previous benchmarks, \shortdatasetname~captures the temporal dependencies inherent in real-world data streams, an aspect often overlooked in traditional evaluations.
Additionally, we propose \shortadvmeminitname, an adversarial memory initialization strategy that enhances the adaptability of TTA methods. By preloading the memory bank with adversarially crafted samples, \shortadvmeminitname~effectively mitigates model forgetfulness and leads to significant performance improvements, particularly in non-\iid~scenarios.
Our findings advocate for a paradigm shift towards benchmarks that more accurately reflect real-world complexities. We emphasize the necessity of robust adaptation strategies capable of handling evolving data distributions, paving the way for the next generation of resilient machine learning models. Future research should explore further enhancements to memory-based adaptation techniques and extend the scope of realistic TTA benchmarks.

\section*{Acknowledgements}
The research reported in this publication was supported by funding from KAUST Center of Excellence on GenAI, under award number 5940.

{\nocite{*}

    \small
    \bibliographystyle{ieeenat_fullname}
    \bibliography{main}
}

\clearpage
\maketitlesupplementary

\section{\shortdatasetname Construction Analysis \& Experimental Setup Details}\
\subsection{Dataset Distribution Overview}
\begin{figure}[h]
    \centering
    \resizebox{\columnwidth}{!}{    {\includegraphics{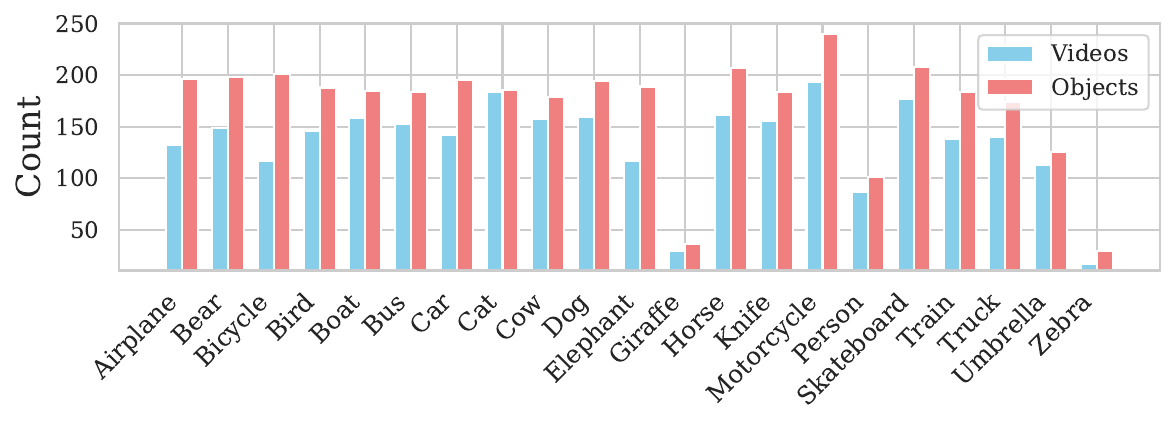}    }
    }
    \caption{
    \textbf{Class distribution of the test set.}
    Here we report a detailed breakdown of the distribution of the 21 object classes in terms of the number of objects and instances.
    }
    \label{fig:class_distrib}
\end{figure}

The dataset distribution of the 21 object classes, as shown in Figure \ref{fig:class_distrib}, presents a varied representation in terms of the number of objects and instances across different classes. The distribution is not uniform, with some classes having a higher number of instances compared to others. This variation provides a diverse range of object occurrences, which can be reflective of different scenarios where certain objects appear more frequently while others are less common. Such a distribution can be valuable in assessing model performance across a broad spectrum of object categories, ensuring that both commonly and less commonly occurring objects are adequately represented in the dataset.

\subsection{Experiment Setup Overview}
Figure ~\ref{fig:setups} provides a visual representation of the two distinct experimental setups used in our study: the frame-wise and tracklet-wise configurations. In the frame-wise setup, a single frame is sampled from each tracklet, ensuring that each object is observed only once. This setup allows for a broad, non-redundant sampling of the dataset, suitable for scenarios where each object instance is considered independently. On the other hand, the tracklet-wise setup processes the frames within each tracklet sequentially, capturing the temporal continuity and variations that occur as the object is observed over time. Both setups are further divided into i.i.d. (independent and identically distributed) and non-i.i.d. settings, providing a comprehensive evaluation framework. The i.i.d. setting assumes that the frames or tracklets are sampled without any dependency, simulating a random observation scenario. Conversely, the non-i.i.d. setting introduces dependencies between samples, reflecting more realistic conditions where observations are not entirely independent, such as in continuous video streams. This dual approach allows for a thorough assessment of model performance under varying assumptions of data distribution and sampling.

\subsection{Controlling the Label Distribution in Streaming Scenarios}\label{sec:label_distrib}

Using a Dirichlet distribution $\textrm{Dir}(\gamma)$ for sampling labels, as introduced in \cite{yuan2023robust}, provides a mechanism for controlling the distribution shift in the label space through the distribution's $\gamma$ parameter. 

Specifically, as $\gamma \rightarrow 0$, we converge to a class-incremental non-\iid setup, while, as $\gamma \rightarrow \infty$, we transition towards a uniform \iid setup. 
We thus extend our experiments by analyzing intermediate stages with $\gamma \in \{10^{-4}, 10^{-1}, 10^{3}\}$, and report the results in Figure~\ref{fig:gamma_study}.

For low values of $\gamma$, the model observes a challenging class-incremental non-\iid scenario. 
In this evolving context, our proposed memory initialization technique proves instrumental in enhancing model performance.
The adversarially initialized memory addresses class bias within the incoming image stream and mitigates degradation in performance.

Conversely, as $\gamma$ increases, the scenario becomes easier, as the model has higher chances of encountering 
images from uniformly-sampled classes. 
In this scenario, the memory samples that were initialized by \shortadvmeminitname are likely to be quickly and completely replaced  with reliable samples from the stream.
This quick replacement is a result of the stream's extreme uniformity, 
which causes any initialization to be inconsequential.
In this context, \shortadvmeminitname neither augments nor diminishes performance.

\subsection{Sensitivity to Batch Size in Streaming Adaptation}

In this section, we explore the impact of batch size on performance.
In particular, we vary the batch size in $\{8, 16, 32\}$, and report our results in Figure~\ref{fig:batch_study}.
As expected, increasing the batch size improves performance, since larger batches provide models with more~(and more diverse) examples on which to adapt. 
When comparing between the \iid and non-\iid setups~(\ie $\gamma \rightarrow \infty$ \textit{vs.} $\gamma \rightarrow 0$), 
we find that methods equipped with \shortadvmeminitname exhibit improved or at least sustained performance across batch sizes. 
However, similar to our observations in previous sections, 
the impact of \shortadvmeminitname is less pronounced in the \iid setting~($\gamma \rightarrow \infty$), indicating that
memory initialization has limited effect in such scenarios.

As mentioned in previous section, we use a default batch size of 64. This batch size ensures that exactly one tracket (64 frames) fits in one forward pass. In previous sections, we experiment with different batch sizes, which are smaller than the default. When the batch size is smaller than the number of frames in a tracklet, we adopt a sequential processing approach, where a tracklet is consecutively processed until all frames are observed. This sequential approach ensures that the entire content of a tracklet is leveraged, potentially enabling the model to adapt effectively to the inherent temporal dependencies and patterns within the data.

However, counterintuitively, we can see from Fig.~\ref{fig:adv_plot_gamma} that smaller batch sizes do not lead to a lower error rate in our experiments. While sequential processing allows for a detailed examination of temporal intricacies within a tracklet, the models, influenced by label distribution, exhibit a degradation in performance with smaller batch sizes. This observation underscores the intricate interplay between batch size, sequential information utilization, and the model's robustness to label distribution.

Importantly, our proposed memory initialization (\shortadvmeminitname) consistently improves or maintains performance compared to standard memory initialization, regardless of batch size. This fact highlights the robustness and effectiveness of \shortadvmeminitname in diverse experimental conditions.

\section{Dynamic Corruption Incorporation: Analysis and Results}

\begin{figure*}[]

    \centering
    \caption{\textbf{Dynamic Severity:} We consider temporal variations in the severity of corruptions, reflecting the realistic scenarios where the impact of corruptions may change over time within a single video clip.}
    \includegraphics[width=\textwidth]{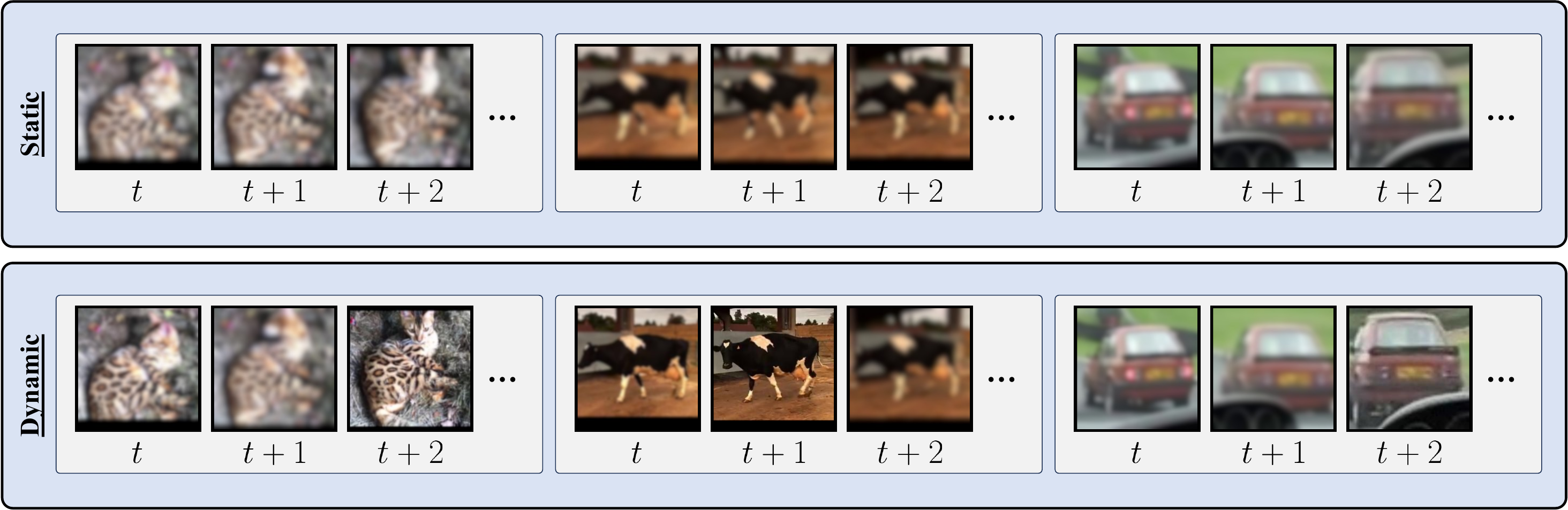}
    \label{fig:dynamic_sev}
    
\end{figure*}
Incorporating dynamic corruptions, as illustrated in Figure~\ref{fig:dynamic_sev}, into our experiments involves the continuous application of corruptions within the tracklet, where the severity level is defined as a function of time. This dynamic approach enables us to precisely control the severity level of each corruption, closely mimicking real-world scenarios. For example, when simulating defocus blur, the dynamic corruption setting introduces fluctuations in focus, alternating between in and out of focus. Similarly, for weather-related corruptions, such as rain, the dynamic application varies the intensity over time, simulating realistic variations in weather conditions within the video clips.
Dynamic corruptions are controlled by the severity function $S(t) = s\:\cdot
\:|\text{sign}(t)|$, where $s$ represents the severity level and $t$ is the index of the frame in a given tracklet. In contrast to static setups, each frame $k_t$ of the $k$-th tracklet experiences variable severity $S(t)$ at time $t$. The severity function can be customized for each corruption type by adjusting the function's parameters (\ie frequency) or additional factors such as random noise.
As depicted in Table \ref{fig:two_tables_dyn}, the deployment of our proposed \shortadvmeminitname in the context of dynamic corruptions exhibits a consistent trend, akin to our findings from the experiments on static corruptions. This observation underscores that the utilization of \shortadvmeminitname consistently again enhances or maintains performance across diverse scenarios.
\begin{figure}[]
    \centering
    \captionsetup{type=table}
    \caption{\textbf{Effect of Adversarial Memory Initialization on TTA performance under Dynamic Severity:} We assess the impact of memory initialization techniques on TTA methods in the dynamic severity setting, where the intensity of corruptions varies within the tracklet. The table presents results for standard~(\xmark) and adversarial~(\cmark) memory initialization~(\shortadvmeminitname).}
    \label{fig:two_tables_dyn}
    \begin{subtable}{0.45\textwidth}
       \centering
        \caption{Tracklet wise \iid}
        \resizebox{\textwidth}{!}{
        \begin{tabular}{lc|ccc|ccc|c}
        \toprule
              &        & \multicolumn{3}{c|}{Noise} &  \multicolumn{3}{c|}{Weather}  \\
        \multirow{-2}{*}{Method} &\multirow{-2}{*}{\shortadvmeminitname}  & gauss. &  shot & impul. &   frost &   fog & brigh. &\multirow{-2}{*}{Avg.}\\
        \midrule
        \multirow{2}{*}{TENT} & \xmark &           89.4 &           89.6 &           91.3 &           92.3 &           91.8 &           92.3 &           91.1 \\
      & \cmark &           68.5 &           69.0 &           70.6 &           81.2 &           73.4 &           56.1 &           69.8 \\\midrule
        \multirow{2}{*}{SHOT-IM} & \xmark &           89.7 &           89.2 &           91.2 &           92.1 &           91.9 &           92.2 &           91.0 \\
      & \cmark &  \textbf{38.4} &  \textbf{38.1} &  \textbf{45.5} &           53.3 &           40.3 &           25.0 &           40.1 \\\midrule
        \multirow{2}{*}{RoTTA} & \xmark &           41.1 &  38.1 &           50.4 &  \textbf{49.4} &  \textbf{38.3} &  \textbf{23.0} &  \textbf{40.0} \\
      & \cmark &           40.7 &           38.3 &           50.0 &           49.6 &           39.1 &           23.4 &           40.2 \\
        \bottomrule
        \bottomrule
        \end{tabular}}
    \end{subtable}
    
    \begin{subtable}{0.45\textwidth}
        \centering
        \caption{Tracklet wise \emph{non}-\iid}
        \resizebox{\textwidth}{!}{
        \begin{tabular}{lc|ccc|ccc|c}
        \toprule
              &        & \multicolumn{3}{c|}{Noise} &  \multicolumn{3}{c|}{Weather}  \\
        \multirow{-2}{*}{Method} &\multirow{-2}{*}{\shortadvmeminitname}  & gauss. &  shot & impul. &   frost &   fog & brigh. &\multirow{-2}{*}{Avg.}\\
        \midrule
 \multirow{2}{*}{TENT}  & \xmark &           89.5 &           89.8 &           91.3 &           92.3 &           91.9 &           92.3 &           91.2 \\
      & \cmark &           78.1 &           78.1 &           82.7 &           87.3 &           84.7 &           78.8 &           81.6 \\\midrule
 \multirow{2}{*}{SHOT-IM} & \xmark &           89.8 &           88.9 &           91.2 &           92.3 &           91.9 &           92.1 &           91.0 \\
      & \cmark &           89.8 &           89.7 &           91.4 &           91.1 &           90.9 &           89.8 &           90.4 \\\midrule
 \multirow{2}{*}{RoTTA}  & \xmark &           61.2 &           62.1 &           68.8 &           75.1 &           66.9 &           59.4 &           65.6 \\
      & \cmark &  \textbf{58.6} &  \textbf{60.3} &  \textbf{67.2} &  \textbf{74.7} &  \textbf{66.1} &  \textbf{51.1} &  \textbf{63.0} \\
        \bottomrule
        \bottomrule
        \end{tabular}}
    \end{subtable}
\end{figure}

\section{Additional Ablations: Vision Transformer (ViT) Experiments}

\begin{figure}[]
    \centering
        \includegraphics[width=\columnwidth]{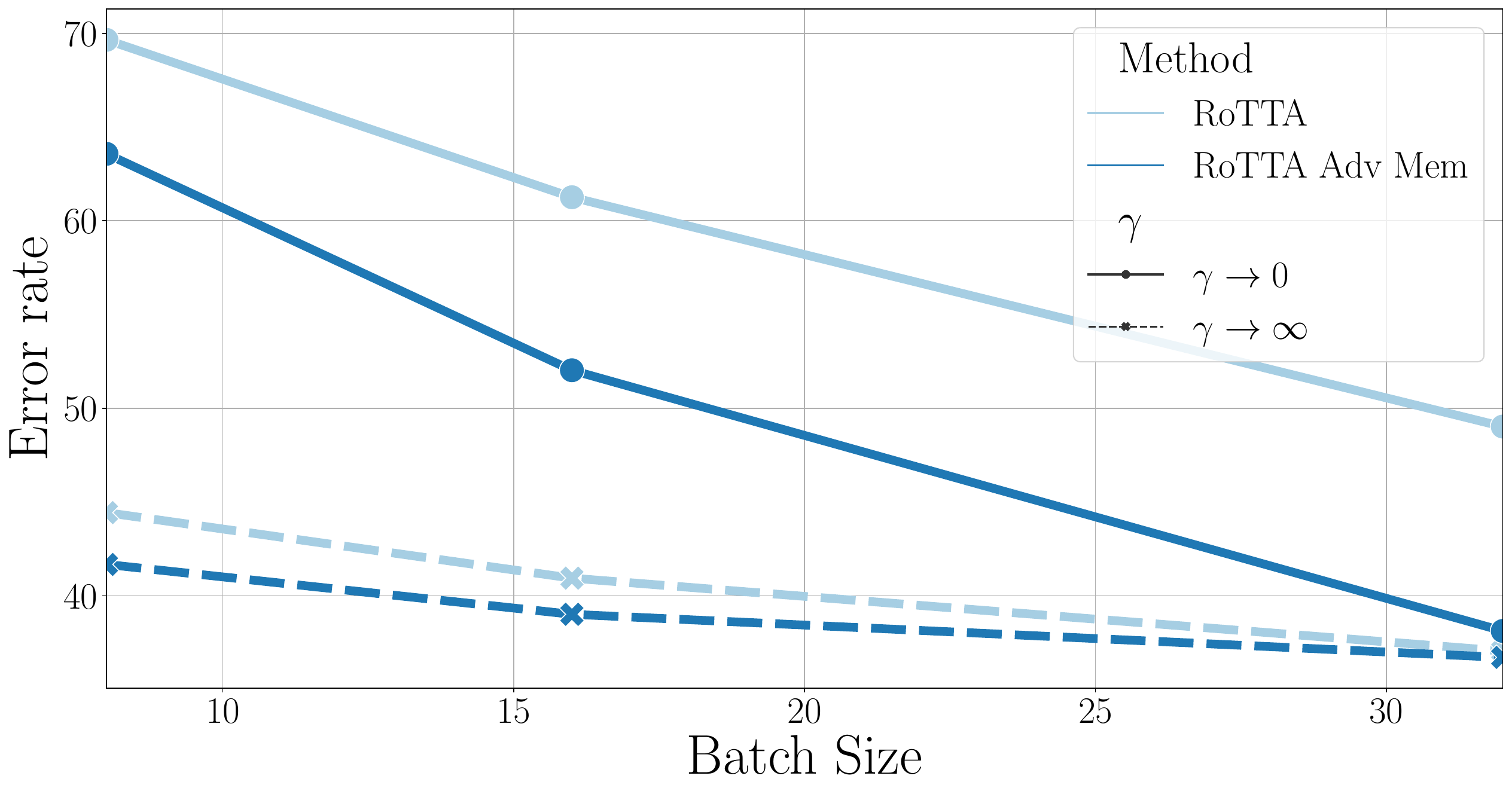}
\caption{\textbf{Error rate as we vary the batch size.}
We study the influence of batch size on method performance. 
Larger batch sizes enhance performance across the board, with our proposed \shortadvmeminitname consistently contributing to further improvements.
All results presented here are for ViT-B-16.}

\label{fig:adv_plot_gamma_vit}
\end{figure}

We extend our experiments to include the Vision Transformer (ViT) architecture. ViT outperforms ResNet-18, even at lower batch sizes, due to its reduced sensitivity to batch size~\cite{sar}. Our ViT experiments focus on the impact of varying batch sizes on method performance (Figure~\ref{fig:adv_plot_gamma_vit}). Larger batch sizes do consistently boost performance, with \shortadvmeminitname adding further improvements. This analysis highlights the adaptability and effectiveness of \shortadvmeminitname.

\section{Evaluation of Memory Initialization using Training Samples}
    \resizebox{\columnwidth}{!}{
        \begin{minipage}{\linewidth}
            \begin{algorithm}[H]
                \caption{TrainMem}
                \label{alg:memory_train_init}
                \begin{algorithmic}
                    \Function{InitializeMemory}{$K, N$}
                        \State Initialize $\mathcal M = \{\}$
                        \While{$|\mathcal M| < N$}
                            \State $y\sim \mathcal U\{1, 2, \dots, K\}$
                                \State $x \sim \mathcal U(\mathcal D\{x|y\})$
                            \State $\mathcal M \gets \mathcal M \cup x$
                            \State $\mathcal D \leftarrow \mathcal D \setminus \{x\}$
                        \EndWhile
                        \State \textbf{return} $\mathcal M$
                    \EndFunction
                \end{algorithmic}
            \end{algorithm}
        \end{minipage}
    }
    \vspace{.5cm}

In contrast to the approach outlined in Algorithm \ref{alg:memory_init}, where adversarial samples are employed, Algorithm \ref{alg:memory_train_init} initializes the memory with images from the training set, denoted by $\mathcal{D}$. Here, $\mathcal{D}$ consists of images and their corresponding labels. 
For any label $y$, $\mathcal{D}\{x|y\}$ represents the subset of $\mathcal{D}$ corresponding to images $x$ labeled as $y$. The initialization process involves uniformly selecting images from $\mathcal{D}$ without replacement until the memory is full.
This procedure, which we refer to as TrainMem, results in a class-wise balanced memory initialization, similar to our adversarial one.

Table~\ref{tab:train_mem} summarizes the results for the tracklet-wise non-\iid setup.
We observe that while initializing the memory with TrainMem, \ie via Algorithm~\ref{alg:memory_train_init}, has positive impact in reducing the error rate of RoTTA, it significantly underperforms our novel \shortadvmeminitname.
For example, TrainMem reduces the error rate against glass blur by 1.7\% when compared to RoTTA.
However, \shortadvmeminitname improves over this naive initialization by over 2\% against the same corruption.

\begin{table*}[]
\caption{\textbf{Tracklet-wise \emph{non}-\iid:} Assessment of TTA methods on \shortdatasetname in a \emph{non}-\iid tracklet context. We contrast standard memory initialization (\xmark), memory initialized with training samples (\cmark\kern-0.65em\small\xmark), and adversarial memory initialization (\cmark).}
\label{tab:train_mem}
\resizebox{\textwidth}{!}{
\begin{tabular}{lc|ccc|ccc|cccc|cccc|c}
\toprule

 & & \multicolumn{3}{c|}{Noise} & \multicolumn{3}{c|}{Blur}                           & \multicolumn{4}{c|}{Weather} & \multicolumn{4}{c|}{Digital} \\
\multirow{-2}{*}{Method} &\multirow{-2}{*}{\shortadvmeminitname}  & gauss. & shot   & impul.  & defoc.& glass & zoom & snow& frost& fog        & brigh.      & contr.      & elast.      & pixel.      & jpeg      & \multirow{-2}{*}{Avg.} \\

\midrule

\multirow{3}{*}{RoTTA} & \xmark &           85.6 &  \textbf{81.6} &           86.5 &           86.9 &           87.3 &           78.9 &           81.7 &           83.5 &           75.0 &           67.2 &           71.8 &           79.7 &           77.1 &           67.4 &           79.3 \\
 & \cmark\kern-0.65em\small\xmark
&  \textbf{83.0} &  \textbf{81.6} &  \textbf{83.1} &           86.3 &           85.6 &           77.3 &           84.4 &  \textbf{79.7} &           75.4 &           71.0 &  \textbf{69.5} &           78.3 &           71.8 &           67.7 &           78.2 \\
      & \cmark &           84.4 &           82.2 &           84.1 &  \textbf{83.0} &  \textbf{83.3} &  \textbf{68.6} &  \textbf{80.2} &           80.5 &  \textbf{74.7} &  \textbf{62.8} &           73.6 &  \textbf{75.2} &  \textbf{61.3} &  \textbf{62.9} &  \textbf{75.5} \\
\bottomrule\bottomrule
\end{tabular}}
\end{table*}

\begin{table*}[!ht]
\caption{
\textbf{Performance Comparison under Frame-wise \iid Assumption.} 
We report average error rates for TTA methods on images from our \shortdatasetname dataset. 
In this setup, all images/frames are shuffled and then independently subjected to corruptions. 
Notably, SHOT-IM outperforms all other methods and across all corruptions, showcasing its robustness against these domain shifts.
\label{tab:fram_iid_A}
}
\vspace{-0.3cm}
\resizebox{\textwidth}{!}{
\begin{tabular}{l|ccc|cccc|cccc|cccc|c}
\toprule

 & \multicolumn{3}{c|}{Noise} & \multicolumn{4}{c|}{Blur}                           & \multicolumn{4}{c|}{Weather} & \multicolumn{4}{c|}{Digital} \\
\multirow{-2}{*}{Method}  & gauss. & shot   & impul.  & defoc.& glass & motion& zoom & snow& frost& fog        & brigh.      & contr.      & elast.      & pixel.      & jpeg      & \multirow{-2}{*}{Avg.} \\

\midrule
Source  &           93.8 &           90.8 &           94.0 &           50.8 &           45.1 &           51.2 &           45.2 &           61.7 &           70.4 &           70.6 &           31.4 &           86.6 &           64.0 &           39.1 &           41.1 &           62.4 \\
AdaBN   &           57.9 &           56.0 &           58.2 &           50.8 &           51.7 &           44.2 &           37.9 &           53.0 &           55.7 &           50.6 &           32.5 &           50.8 &           37.8 &           29.0 &           36.8 &           46.9 \\
SHOT-IM &  \textbf{47.5} &  \textbf{45.5} &  \textbf{47.1} &  \textbf{43.6} &  \textbf{43.2} &  \textbf{36.4} &  \textbf{31.2} &  \textbf{47.1} &  \textbf{50.2} &  \textbf{41.7} &  \textbf{27.8} &  \textbf{44.5} &  \textbf{29.6} &  \textbf{24.6} &  \textbf{30.0} &  \textbf{39.3} \\
TENT    &           57.9 &           55.8 &           58.4 &           50.6 &           51.7 &           44.2 &           37.8 &           52.9 &           55.6 &           50.4 &           32.5 &           50.4 &           37.7 &           28.9 &           36.8 &           46.8 \\
SAR     &           57.6 &           55.3 &           59.1 &           50.4 &           51.3 &           43.7 &           37.6 &           52.8 &           55.4 &           50.1 &           32.2 &           49.4 &           37.6 &           28.6 &           36.3 &           46.5 \\
CoTTA   &           57.9 &           55.8 &           58.6 &           50.3 &           51.3 &           43.8 &           37.6 &           53.0 &           55.8 &           50.4 &           32.3 &           50.4 &           37.8 &           28.9 &           36.7 &           46.7 \\
ETA     &           57.9 &           54.6 &           59.0 &           50.5 &           51.6 &           44.0 &           37.5 &           53.1 &           55.7 &           50.7 &           32.4 &           50.9 &           37.4 &           28.8 &           36.7 &           46.7 \\
EATA    &           57.9 &           55.1 &           58.1 &           50.2 &           51.9 &           43.4 &           38.0 &           54.4 &           56.2 &           50.9 &           31.9 &           50.3 &           38.8 &           28.5 &           36.7 &           46.8 \\
RoTTA   &           69.4 &           64.0 &           71.2 &           54.4 &           55.0 &           50.6 &           42.4 &           60.0 &           62.0 &           59.6 &           34.3 &           64.3 &           43.3 &           31.1 &           38.7 &           53.4 \\
\toprule
\bottomrule
\end{tabular}
}
\end{table*}
\begin{table*}[!ht]
\caption{\textbf{Tracklet-wise \iid Evaluation with and without Memory.} 
We report average error rates for TTA methods on our \shortdatasetname dataset under the tracklet-wise \iid scenario (\ie entire tracklets are sampled \iid). 
The results are grouped to reflect the presence (\cmark) or absence (\xmark) of memory during adaptation. 
RoTTA shows notable robustness by using memory, significantly outperforming the non-memory variants across various corruption types, as demonstrated by the marked difference in error rate.
\label{tab:tracklet_iid_mem_A}}
\vspace{-0.3cm}
\resizebox{\textwidth}{!}{
\begin{tabular}{lc|ccc|ccc|cccc|cccc|c}

\toprule
 & & \multicolumn{3}{c|}{Noise} & \multicolumn{3}{c|}{Blur}                           & \multicolumn{4}{c|}{Weather} & \multicolumn{4}{c|}{Digital} \\
\multirow{-2}{*}{Method} &\multirow{-2}{*}{Memory}  & gauss. & shot   & impul.  & defoc.& glass & zoom & snow& frost& fog        & brigh.      & contr.      & elast.      & pixel.      & jpeg      & \multirow{-2}{*}{Avg.} \\

\midrule
TENT & \xmark &           94.3 &           94.3 &           94.4 &           94.5 &           94.6 &           94.1 &           94.1 &           94.5 &           93.4 &           93.1 &           93.7 &           94.0 &           93.5 &           93.7 &           94.0 \\
      & \cmark &           94.2 &           94.2 &           94.3 &           94.5 &           94.5 &           93.9 &           93.7 &           93.7 &           93.2 &           93.0 &           92.8 &           93.8 &           93.4 &           93.5 &           93.8 \\\midrule
SHOT-IM & \xmark &           94.8 &           94.7 &           94.8 &           94.8 &           94.7 &           94.8 &           94.7 &           94.8 &           94.6 &           94.4 &           94.7 &           94.9 &           94.7 &           94.8 &           94.7 \\
      & \cmark &           93.8 &           93.9 &           93.9 &           94.2 &           94.3 &           93.7 &           93.3 &           93.0 &           93.0 &           92.8 &           92.1 &           93.5 &           93.4 &           93.2 &           93.4 \\\midrule
RoTTA & \cmark &  \textbf{68.0} &  \textbf{62.4} &  \textbf{68.6} &  \textbf{58.5} &  \textbf{56.7} &  \textbf{40.9} &  \textbf{56.6} &  \textbf{60.7} &  \textbf{52.2} &  \textbf{30.2} &  \textbf{60.7} &  \textbf{39.7} &  \textbf{27.8} &  \textbf{35.1} &  \textbf{51.3} \\
\bottomrule
\end{tabular}
}
\end{table*}
\begin{table*}[!ht]
\caption{
\textbf{Tracklet-wise \textit{non}-\iid Evaluation with and without Memory Banks.}
We report the performance of TTA methods when tracklets are non-\iid, \ie tracklets are sampled such that their labels display correlation in time~(by following a Dirichlet distribution).
We further examine whether using a memory bank (\cmark) influences outcomes. 
RoTTA with memory exhibits the lowest error rates, indicating proficiency against non-\iid data. 
Without memory (\xmark), all methods experience worse error rates, underscoring the impact of memory in adapting to complex data streams.
\label{tab:tracklet_non_iid_mem_A}}
\resizebox{\textwidth}{!}{

\begin{tabular}{lc|ccc|ccc|cccc|cccc|c}
\toprule
 & & \multicolumn{3}{c|}{Noise} & \multicolumn{3}{c|}{Blur}                           & \multicolumn{4}{c|}{Weather} & \multicolumn{4}{c|}{Digital} \\
\multirow{-2}{*}{Method} &\multirow{-2}{*}{Memory}  & gauss. & shot   & impul.  & defoc.& glass & zoom & snow& frost& fog        & brigh.      & contr.      & elast.      & pixel.      & jpeg      & \multirow{-2}{*}{Avg.} \\

\midrule
TENT & \xmark &           94.3 &           94.3 &           94.4 &           94.5 &           94.6 &           94.0 &           94.0 &           95.0 &           93.5 &           93.2 &           93.5 &           94.0 &           93.5 &           93.7 &           94.0 \\
      & \cmark &           94.2 &           94.3 &           94.3 &           94.5 &           94.5 &           94.0 &           93.7 &           93.6 &           93.4 &           93.0 &           92.9 &           93.8 &           93.4 &           93.6 &           93.8 \\
\midrule
SHOT-IM & \xmark &           95.1 &           94.9 &           95.1 &           95.1 &           95.1 &           95.7 &           94.9 &           95.2 &           94.8 &           95.0 &           95.0 &           94.8 &           95.3 &           95.2 &           95.1 \\
      & \cmark &           93.9 &           93.8 &           93.9 &           94.3 &           94.3 &           93.7 &           93.4 &           93.6 &           93.2 &           92.7 &           92.5 &           93.6 &           93.5 &           93.2 &           93.6 \\
\midrule
RoTTA & \cmark &  \textbf{85.6} &  \textbf{81.6} &  \textbf{86.5} &  \textbf{86.9} &  \textbf{87.3} &  \textbf{78.9} &  \textbf{81.7} &  \textbf{83.5} &  \textbf{75.0} &  \textbf{67.2} &  \textbf{71.8} &  \textbf{79.7} &  \textbf{77.1} &  \textbf{67.4} &  \textbf{79.3} \\

\bottomrule\bottomrule
\end{tabular}
}
\end{table*}
\section{Visualizing Adversarial Examples for \shortadvmeminitname Initialization}
\begin{figure}[H]
    \centering
    \subcaptionbox{Airplane}
    {\begin{subfigure}[a]{0.4\columnwidth}
        \centering
        \includegraphics[width=0.8\columnwidth]{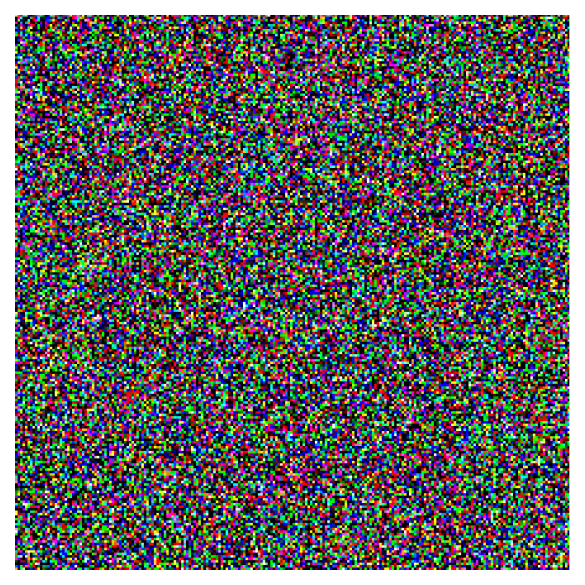}
    \end{subfigure}}
    \subcaptionbox{Bear}
    {\begin{subfigure}[a]{0.4\columnwidth}
        \centering
        \includegraphics[width=0.8\columnwidth]{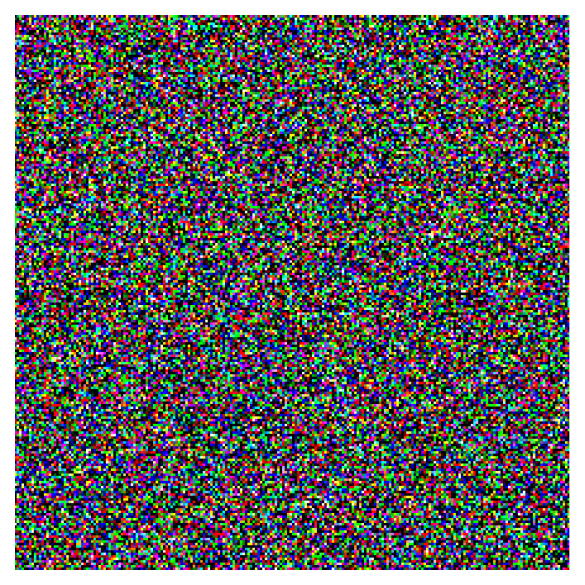}
    \end{subfigure}} 
    \subcaptionbox{Bicycle}
    {\begin{subfigure}[a]{0.4\columnwidth}
        \centering
        \includegraphics[width=0.8\columnwidth]{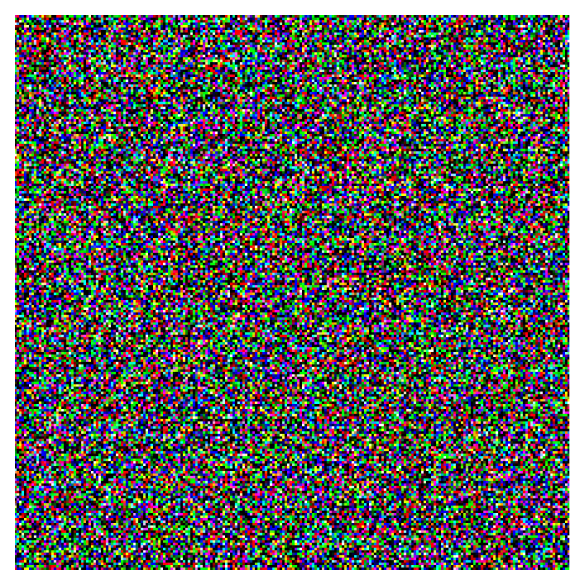}
    \end{subfigure}} 
    \subcaptionbox{Bird}
    {\begin{subfigure}[a]{0.4\columnwidth}
        \centering
        \includegraphics[width=0.8\columnwidth]{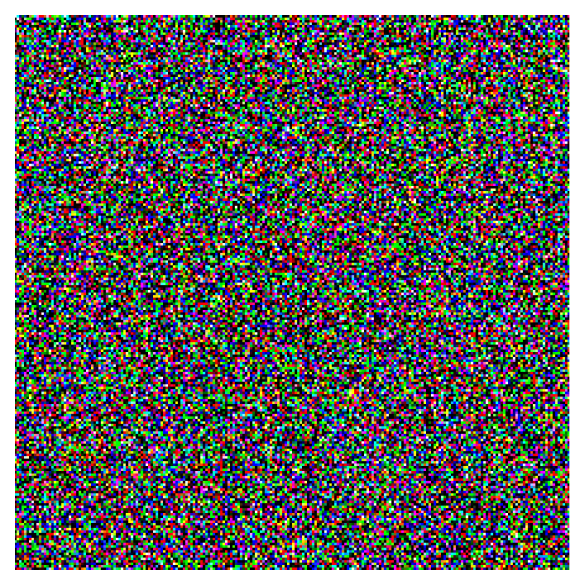}
    \end{subfigure}}
  
    \caption{Visualizations of selected adversarial examples used for initializing the memory bank in \shortadvmeminitname.}
    \label{fig:adv_samples}
\end{figure}

In this section, we present visualizations of adversarial examples used for initializing \shortadvmeminitname. These examples are generated during the memory bank initialization process. For details on the creation of adversarial examples, please refer to \shortadvmeminitname sections. Figure~\ref{fig:adv_samples} showcases a selection of these adversarial examples.

\section{Supplementary Table Details}
\vspace{-.25cm}
This appendix we present the expanded versions of the tables from the main paper, maintaining the same titles for consistency. These tables contain additional data and detailed results. The supplementary information includes detailed breakdowns and results. These expanded tables are intended to offer a comprehensive reference for readers seeking further insights and details related to the study.

\end{document}